# When Does Geometric View Synthesis Help Wine Label Retrieval? A Public One-Shot Benchmark Across Self-Supervised and Vision–Language Backbones

**Yueh-Cheng Huang[1]**

[1]Department of Computer Science and Information Engineering, National Dong Hwa University, Hualien 974301, Taiwan

Corresponding author: Yueh-Cheng Huang (e-mail: yuehch@gms.ndhu.edu.tw).

This research received no external funding.

ABSTRACT Geometric view synthesis can expand a single wine-label photograph into a training set, but its value with pretrained image encoders is unclear. We study this on a public WineSensed-derived benchmark of 1,000 classes, one enrollment photograph per class, and 4,295 real queries. With the earlier DINO vision transformer (ViT-S/16) recipe, geometric views raise top-1 accuracy from 34.1% to 62.6–63.7%, about three times the gain from two-dimensional (2D) augmentation. Frozen SigLIP 2-B already reaches 94.7%. A linear head over its frozen features gains 1.2–1.3 percentage points with the two geometric pipelines localized by the Segment Anything Model (SAM), while the other pipelines gain an inconclusive 0.3–0.6 points. Low-rank adaptation (LoRA) and validation-selected full fine-tuning show no clear gain within the reported confidence intervals; fixed-budget full fine-tuning loses 9–24 points. SAM localization supplies all six views for 99% of sources, compared with 43% for the edge-based front end. Recognition differences between the two cylinder constructions depend on the training recipe and are confounded by their crop and canvas conventions. Rendered-cylinder tests show different responses to source tilt, but an uncalibrated rim-ratio proxy establishes no corresponding trend in recognition on real photographs. An author-confirmed audit of 50 residual errors identifies 21 query–enrollment appearance mismatches, without establishing an irreducible error rate. These results support geometric synthesis for the tested self-supervised recipe and a smaller benefit through frozen-feature adaptation of the text-supervised encoder.

INDEX TERMS Benchmark testing, data augmentation, image retrieval, one-shot learning, vision–language models, view synthesis, wine label recognition.

## I. INTRODUCTION

Identifying a wine from a phone photograph of its label is a routine feature of consumer applications and a recurring subject of academic work. As a retrieval problem it has an unusual profile: the number of classes is large, the discriminative content is mostly printed text and layout, and in practice a new label may be represented by a single reference photograph. The academic systems with the highest reported accuracy [1], [2], [3] share one design. A convolutional network trained on a large proprietary collection (about 550,000 images in [1]) narrows the search to a producer, and hand-crafted local features then select the product within that producer; sub-brand accuracy in this line of work is 82–84%. The private datasets limit independent reproduction and direct comparison of the reported results.

A second line of work, to which our earlier papers [4], [5] belong, addresses the shortage of training images per label. A label is printed on a cylinder, so a single photograph can be re-projected to other viewpoints once the cylinder geometry has been recovered from the image, and the re-projected views can be used as training data. In [4] we estimated the label's rims and the vanishing point of its edges and resampled the surface through a cross-ratio construction; a metric-learning model trained on the synthesized views improved one-shot top-1 accuracy by about 14 points over conventional two-dimensional (2D) augmentation in that setting. In [5] the edge-based label localization of that pipeline was replaced by the Segment Anything Model (SAM) [6]. Independently, Li et al. [7] proposed a cylinder model with three viewpoint motions that takes the input photograph to be frontal, and reported a gain of 4.1 points in main-brand accuracy over image copying on a proprietary collection, together with a comparison against generative augmentation. The two constructions differ in what they assume about the input, and they have not been compared on common data.

Both lines predate the current practice of starting retrieval from a large pretrained encoder. Frozen self-supervised encoders such as DINO [8] and DINOv2 [9] and text-supervised encoders such as SigLIP 2 [10] provide strong instance-level retrieval with no task-specific training, and a recent large-scale benchmark [11] reports that such encoders, sometimes with a light adaptation layer, are competitive with or better than domain-trained models. Label recognition should be a favorable case for text-supervised encoders in particular, because what distinguishes one label from another is largely printed text and layout. This raises a question the earlier work could not ask: once a strong pretrained encoder is the starting point, does geometric view synthesis still contribute anything, and if so, for which backbone and through which form of adaptation?

This paper answers that question on a public benchmark. The benchmark is derived from the WineSensed release [12], which contains user-uploaded label photographs from the Vivino platform, and consists of 1,000 label classes with one enrollment photograph each and 4,295 real query photographs, with a nested 100-class subset for ablations and a separate 1,384-class split used only for development and model selection. On it we ask three things: how accurate a frozen pretrained embedding is at one-shot wine-label retrieval; whether geometric view synthesis, used either to adapt the encoder or to enrich the gallery, improves on that embedding, and how the answer depends on the backbone and on how much of the encoder the adaptation is allowed to change; and what kinds of errors remain. Fig. 1 summarizes the design.

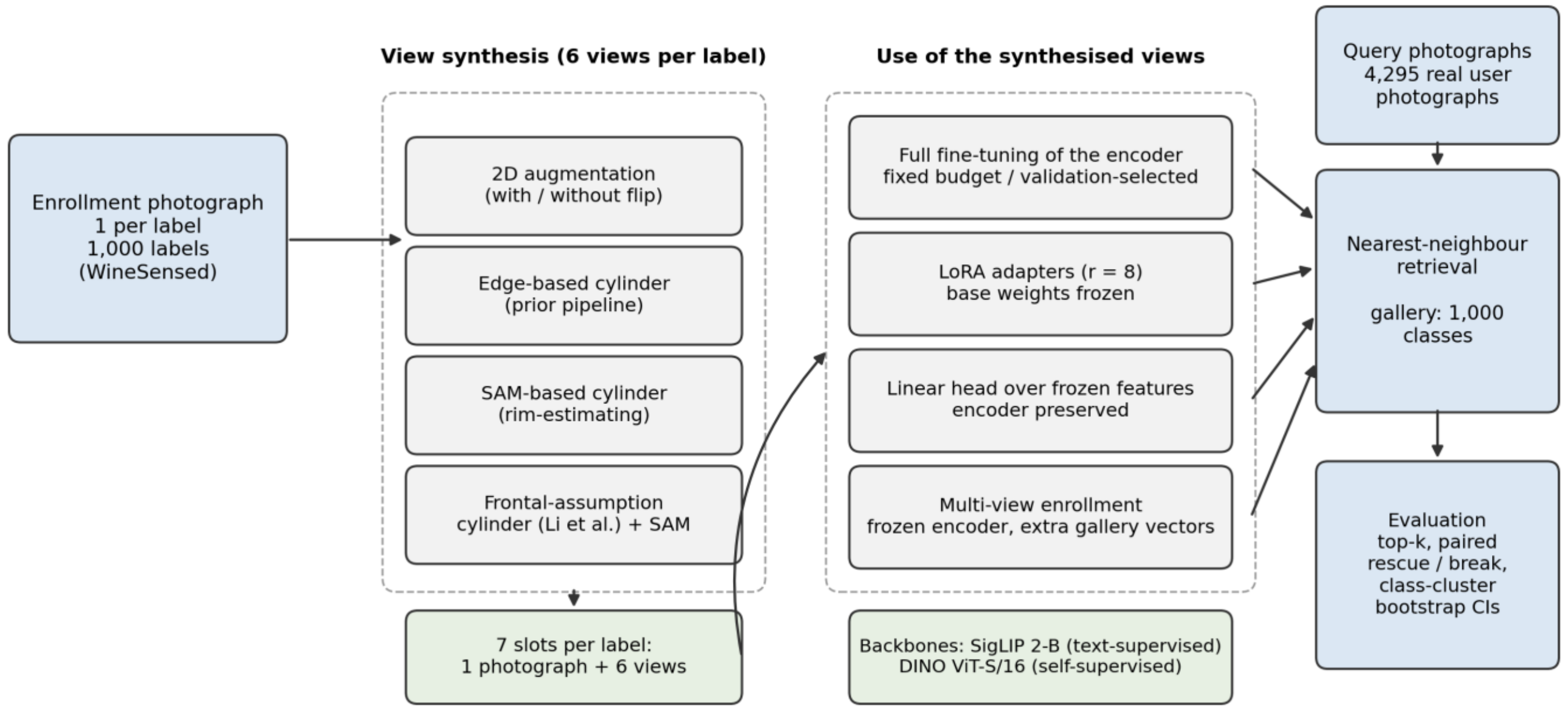

**FIGURE 1. Overview of the study. One enrollment photograph per label is passed through four view-synthesis pipelines (2D augmentation with and without flipping, the edge-based and SAM-based cylindrical pipelines, and the frontal-assumption construction of [7] with SAM localization). The views are used either to adapt an encoder (full fine-tuning, low-rank adaptation (LoRA) adapters, or a linear head over frozen features) or as additional enrollment vectors; retrieval against the 1,000-label gallery is evaluated on 4,295 real user photographs with two backbones, and the frozen encoder without training is the reference for every comparison. In the figure, ViT refers to vision transformer and CIs to confidence intervals.**

The short answer is that there are two regimes. When the backbone is a self-supervised encoder with no strong prior for printed labels, geometric synthesis is decisive: it adds about 30 points of top-1 accuracy over the frozen model, roughly three times the gain from 2D augmentation, consistent with the direction of the gains reported on private data in [4] and [7]. When the backbone is a text-supervised encoder that already recognizes 94.7% of the queries without any training, the views are worth about one point, and only through an adaptation that leaves the encoder untouched. The outcome depends on the tested adaptation recipe: a linear head over frozen features gains 1.2–1.3 points with the two SAM-localized geometric pipelines, LoRA adapters gain nothing the confidence intervals support, and full fine-tuning at a fixed budget loses 9–24 points. The clearest front-end benefit is coverage: replacing the edge-based localizer by SAM raises the share of sources supplying all six views from 43% to 99%. Differences between the cylinder constructions depend on the adaptation recipe and cannot be isolated from their crop and canvas conventions. In an audited sample, 21 of 50 residual errors involve visibly different query and enrollment labels; this observation does not determine the maximum achievable accuracy.

The contributions are as follows.

1) A public one-shot benchmark for wine-label retrieval with prospectively frozen splits, released identifier lists and hashes, and evaluation code. Images are not redistributed, in keeping with the source license, but the benchmark can be rebuilt from the official release. The wine-label retrieval studies reviewed in Section II-A do not provide this evaluation protocol on a shared public benchmark.

2) A measurement of frozen backbones on this benchmark: a frozen SigLIP 2-B embedding reaches 94.7% top-1 with no training, and a frozen DINO ViT-S/16 reaches 34.1%.

3) A controlled comparison of four view-synthesis families (2D augmentation, with an additional control without horizontal flipping, the edge-based cylindrical pipeline of [4], a SAM-based cylindrical pipeline, and a frontal-assumption pipeline after [7]) on both backbones, under matched budgets, three seeds, and class-cluster confidence intervals, and under adaptation recipes that update different fractions of the encoder, from full fine-tuning to a linear head over frozen features, with multi-view enrollment as a training-free alternative.

4) A comparison of the front end and cylinder construction, through a 2 × 2 comparison of the two components, a ground-truth cylinder renderer that shows where the two constructions differ, and an uncalibrated pose proxy that characterizes the enrollment photographs of the benchmark.

5) A measurement of how far synthesized views lie from real photographs in the SigLIP 2 feature space, and a fixed-seed human audit of 50 of the 226 residual errors.

This paper builds on our conference papers [4], [5]. The synthesis geometry is inherited from [4] and the SAM front end from [5], and both are summarized with citation in Section III; the benchmark, every experiment reported here, and the text are new; results of the conference papers are cited with attribution, but none is reported as a result of this paper.

## II. RELATED WORK

### A. WINE-LABEL RECOGNITION

Published wine-label recognizers treat the task as retrieval of a query photograph against a set of reference label images, and the most accurate of them follow the consecutive search-and-match design introduced by Li, Yang, and Ma [1]: a convolutional neural network (CNN) trained to classify the main brand narrows the search, and scale-invariant feature transform (SIFT) matching with random sample consensus (RANSAC) verification and term frequency–inverse document frequency (TF-IDF) weighting selects the sub-brand. The same authors replaced SIFT by speeded-up robust features (SURF) and reported 82.3% sub-brand accuracy [2], and later distributed the CNN over branches trained on subsets partitioned by class frequency, reaching 84.0% at 1.36 s per query [3]. All three papers use a collection of about 548,000 images from a commercial partner, with 17,328 main brands and 260,579 sub-brands, which is not available. A text-based alternative reads the label with optical character recognition (OCR) and searches a structured database of names [13]; it was evaluated on 45 bottles. Commercial applications exist but publish no methodology.

### B. VIEW SYNTHESIS AND AUGMENTATION FOR LABELS

A wine label is printed on a cylinder, so a single frontal photograph can be re-projected to other viewpoints if the cylinder geometry is recovered from the image. Two constructions have been proposed for this, and they differ in what they assume about the input photograph. Li et al. [7] model the label as a cylinder of known radius seen from a viewpoint at a known distance, derive closed-form pixel correspondences for lateral, radial, and vertical viewpoint motion, and remove the black borders that the re-projection leaves; the input is taken to be a frontal projection, and a fully convolutional network trained on 5,000 annotated images segments the label region. On their collection the synthesized views raised main-brand accuracy from 0.716 (image copying) to 0.757, above DAGAN (0.719) and Fast AutoAugment (0.744). Our earlier pipeline [4] does not assume a frontal input: it detects the upper and lower rims of the label in the image, estimates the left and right edges, takes their intersection as the vanishing point of the cylinder axis, and resamples the pixels along each longitudinal line onto a cylinder of chosen pose using the cross-ratio, which is invariant under projection. Label unwrapping with similar geometry appears in engineering write-ups and industrial vision software [14], usually without an evaluation. Generative alternatives, such as the generative adversarial network (GAN) that Tonioni and Di Stefano [15] use to bridge studio and shelf images, need training data of their own; [7] contains the only direct comparison between geometric and generative augmentation for labels that we are aware of, and generative augmentation is outside the scope of this paper.

### C. RETRIEVAL WITH PRETRAINED EMBEDDINGS

Instance-level retrieval with frozen features from self-supervised transformers [8], [9] and from image–text models [10] has become standard practice. The ILIAS benchmark [11] finds that models fine-tuned on one product domain generalize poorly to others, that a linear adaptation of image–text models helps, and that local descriptors used for re-ranking remain valuable under clutter; learned re-rankers such as AMES [16] refine the candidate list that a global embedding returns. In fine-grained product recognition, scene text extracted by OCR has been combined with visual features [17], [18], [19], and a recent study of grocery products reports that OCR text alone is weaker than the image but helps on near-identical packaging [20]. These findings shape the controls used here: the main results are image-only, a local-feature re-ranker and a simple OCR baseline are evaluated only as controls, and catalog-to-real retrieval with hard-negative mining [21], a related setting, is not pursued because the benchmark provides one enrollment image per class.

## III. VIEW-SYNTHESIS PIPELINES

Four pipelines produce six training or enrollment views from each source photograph: three geometric pipelines and a 2D

control. The three geometric pipelines provide two pairwise comparisons of their localization and construction components. The first two share the cylinder construction of [4] and differ in how the label is located in the photograph (an edge detector or SAM); the last two share the SAM front end and differ in the construction (ours or that of [7]). Section VI-D examines these comparisons while accounting for differences in fallback coverage and rendering conventions. Fig. 2 shows examples and failures.

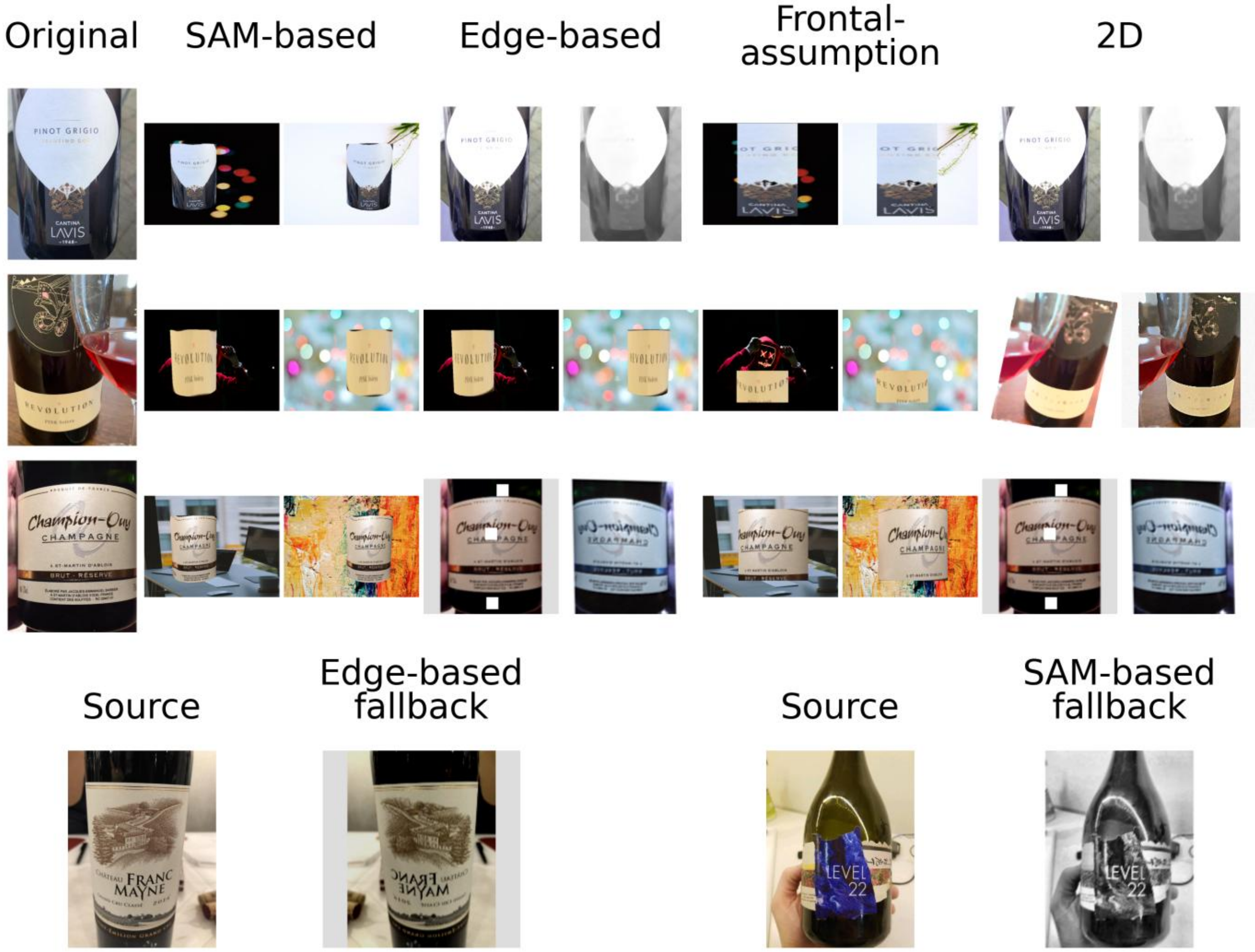

**FIGURE 2. Synthesis examples. Three sources drawn with a fixed seed (top three rows): the original photograph and the first two synthesized views from each pipeline. Bottom row: an edge-based failure and a SAM failure with the 2D fallback image that replaced the slot. The edge-based views in the first and third rows are also 2D fallbacks. Edge-based and SAM-based denote the three-dimensional (3D) pipelines based on [4] and [5], respectively; Frontal-assumption denotes the construction adapted from Li et al. [7] with SAM localization.**

### A. CYLINDRICAL MODEL AND CROSS-RATIO RESAMPLING

The construction is that of [4] and is summarized here so that the paper is self-contained. Let the label be a patch on a right circular cylinder whose axis is vertical in the world. In the image, the upper and lower rims of the label are arcs of ellipses, and the left and right edges of the label, which are generators of the cylinder, meet at the vanishing point D of the axis direction. For a pixel A on the wider rim, the corresponding pixel C on the other rim lies where the line AD crosses that rim, and the segment AC is the image of one longitudinal line of the label. A cylinder of the desired pose is placed in a virtual scene and projected, which gives a target segment A′C′ for each line, and points along AC are mapped to A′C′ by requiring the cross-ratio (A′C′·B′D′)/(A′D′·B′C′) to equal the cross-ratio of the corresponding points on AC. Because the mapping along a line depends only on the positions of its endpoints and of the vanishing point, it is computed exactly for the central line and applied to the other lines by scaling, as in [5]; Section III-E measures what this simplification costs. The angular extent of the visible surface is fixed at 160 degrees, a value chosen in [5] from the mean visible angle of 320 rendered cylinder poses.

### B. FRONT ENDS

The pipelines differ in how the rims and edges are found, and this is the step at which synthesis can fail.

*Edge-based localization.* This is the front end of [4], run from its original implementation with crash fixes only. The image is converted to gray, pixels with a strong vertical gradient are collected, blocks containing enough such pixels are chained by gradient sign, non-maximum suppression thins the chains, and a curve is fitted to each rim; the left and right edges are the common external tangents of the two fitted ellipses. The detector was designed for photographs of higher resolution than those of the benchmark, and it serves as the reference against which the SAM front end is measured.

*SAM-based localization.* A text prompt to Grounding DINO [22] locates the label and SAM [6] returns its mask. Working outward from the center column of the mask, the topmost and bottommost mask pixels in each column form coarse upper and lower rims, and a column is dropped when its vertical jump from its neighbor exceeds a threshold. An ellipse is fitted to each coarse rim, or a parabola when the fitted ellipse is degenerate or much shorter than the rim. The left and right edges join the endpoints of the two rims, and their intersection is the vanishing point. Failures are detected from the mask area, the rim length, the fit residual, and the position of the vanishing point.

*Frontal-assumption construction adapted from [7], with SAM localization.* The third pipeline applies the cylinder construction of [7] using the same SAM localization as the second. Their rendering conventions also differ, as detailed below. Following [7], the crop (resized to a maximum side of 256 px) is treated as a frontal projection of a cylinder of radius $r_0$ seen from distance $r_1$, and new views are produced for a lateral rotation of ±20°, a radial move to ×0.75 and ×1.25 of the distance, and a vertical move of ±15°. Since [7] does not give $r_1/r_0$, we chose 10 from {3, 5, 10} on the development split before any formal experiment, and the black borders left by the re-projection are removed as in [7]. Two departures from [7] should be noted. Its label region is segmented by a fully convolutional network trained on 5,000 images from a 9,000-image annotated segmentation dataset, whereas here SAM performs that step, so our results say nothing about that network; and in [7] the construction augments a classifier trained on many images per label, whereas here it produces six views from a single photograph, like the other pipelines. What the construction does not do is estimate the pose of the input: a photograph taken from above or below the label is re-projected as if it were frontal, and Section III-E measures the effect of this assumption. We refer to it as the frontal-assumption construction and to ours, which estimates the rims and the vanishing point, as the rim-estimating construction.

Views from all three geometric pipelines are composited onto a background at the position of the label in the source photograph. For a given source and view index, every pipeline uses the same background image and placement, so the pipelines differ only in the label region. Within that region, however, the two constructions are not on a common footing: the rim-estimating construction renders the label onto its native 642 × 482 canvas, whereas the frontal-assumption construction operates on the crop resized to a maximum side of 256 px, so the scale and projection conventions of the label region differ between them as well as the geometry. This is a limit on the fairness of the construction comparison that the experiments do not control.

### C. 2D CONTROL

The 2D pipeline is the augmentation used for the 2D baseline of [4], applied to the whole enrollment photograph with its background, six times per source with a fixed seed. Each view passes through the following stages in order: horizontal flip (p = 0.5); one of perspective warp, affine transform (rotation ±15°, translation ±5%, shear ±7°), or grid distortion (p = 0.5); one of four blurs (p = 0.5); one of six color operations dominated by color jitter (p = 0.8); one of channel dropout, conversion to gray, or channel shuffle (p = 0.2); and coarse dropout or Gaussian noise (p = 0.25). Having no geometric model, it cannot fail on a source and needs no fallback.

Horizontal flipping mirrors the printed text (Fig. 2), while conversion to gray removes color information, and [7] reports that flipping lowers accuracy on their collection. We keep the pipeline as in [4] as the primary control and add a second 2D bank, identical except that the horizontal flip is removed, to check whether the flip is responsible for what the control does (Sections VI-B and VI-C). Removing an operation changes the random draws of the operations that follow, so 5,799 of the 6,000 views of the flip-free bank differ from the original bank in more than the flip; the two banks compare the pipeline with and without flipping, not image by image.

### D. SLOTS AND FALLBACK

A geometric pipeline can fail on a source photograph, and if failed sources were simply dropped, the pipelines would be trained on different classes and could not be compared. Each source therefore has one original image and six synthesis slots, 7,000 slots per pipeline for the 1,000-class benchmark, and when a geometric pipeline fails on a slot, the slot is filled with a fixed 2D-augmented image of the same source and view index. Every class keeps the same number of training images, no source is dropped, and the number of replaced slots is reported with every result. Over the 1,000 sources, the edge-based pipeline synthesized 2,588 of 6,000 views natively (431 sources with all six views), the SAM-based pipeline 5,958 (993 sources), and the frontal-assumption pipeline 5,988 (998 sources).

### E. GEOMETRIC VALIDATION

Recognition accuracy cannot tell whether the synthesized geometry is right, so the implementation was checked against a ray-cast renderer of a textured cylinder with a pinhole camera. A known texture is rendered at pose A, the pipeline synthesizes pose B from that image using the ground-truth mask, and the result is compared with the direct rendering of pose B: for each output pixel, the input pixel actually sampled is traced back to the cylinder surface and re-projected, and its distance from the true position is recorded in output pixels. On 33 pose pairs (3 source poses × 11 targets) at 642 × 482 pixels, the test exposed an error in the handling of near-vertical lines, with a maximum displacement of 260 px; after correction the maximum is 18.6 px and the mean over poses 4.1 px (Table S1). The same test measures the central-line simplification of Section III-A. On the 22 pose pairs where all three sampling variants are defined, exact per-line projection gives a mean displacement of 0.54 px, the central-line cross-ratio simplification 0.60 px, and equidistant sampling, which ignores perspective, 4.0 px with a maximum of 15.3 px. The simplification costs almost nothing; ignoring perspective does not.

Two qualifications apply. The ground-truth mask replaces SAM in this test, so the numbers measure the geometry and sampling, not the segmentation. And the remaining 4 px mean displacement of the full pipeline reflects the approximate treatment of the visible angle and the fitted rims, which is inherited from [4]; the method was not changed beyond the numerical correction.

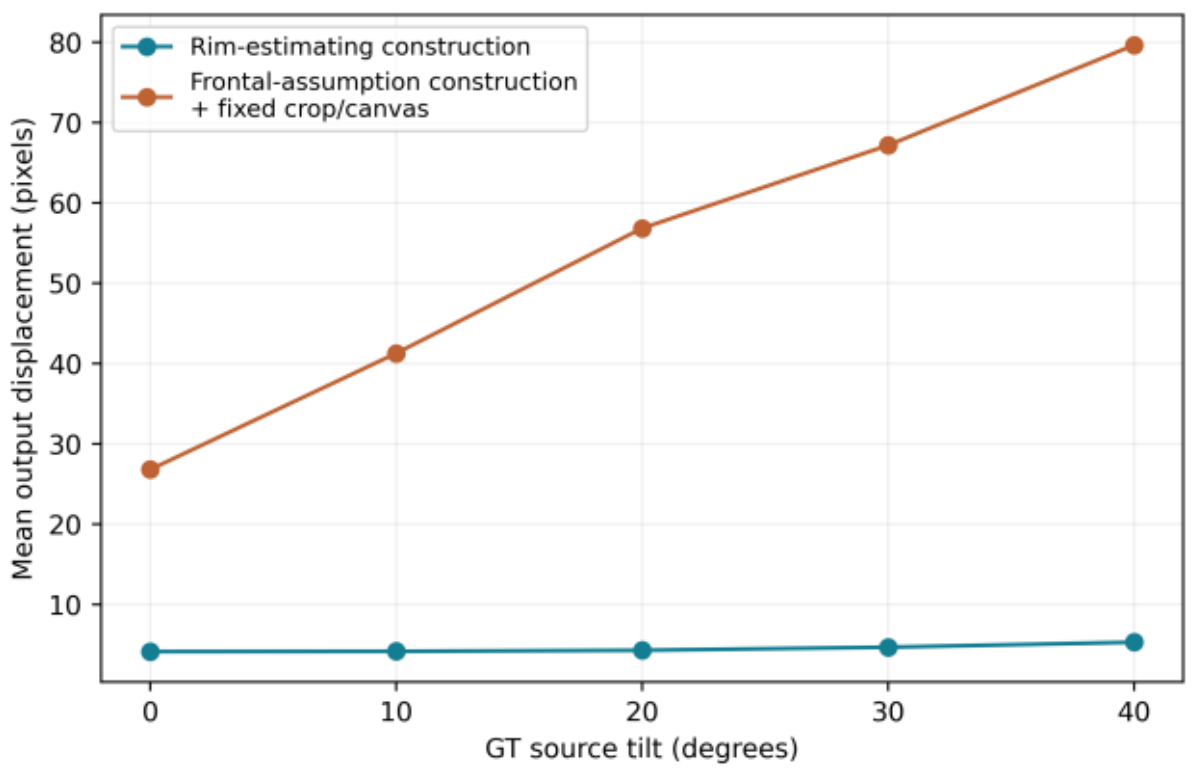

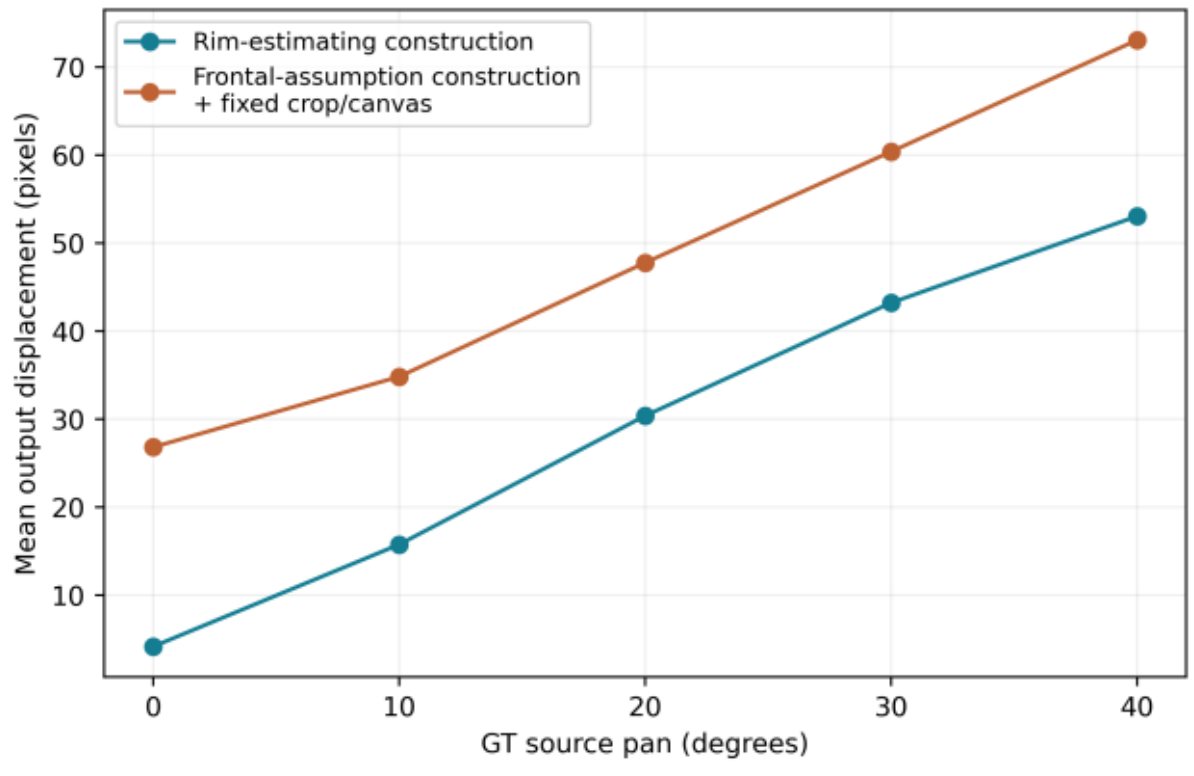

**FIGURE 3. Re-projection displacement of the two constructions against the pose of the source on the rendered cylinder (Table S2): (a) tilt toward the camera, (b) rotation about the cylinder axis. Mean over the 11 target poses; the frontal-assumption construction carries a fixed offset from its crop and canvas adaptation, so the curves are compared in slope.**

*Response to the pose of the input.* The property that distinguishes the two cylindrical constructions is how they respond to a tilted input, and the renderer can measure it directly. The source cylinder was tilted toward the camera (rotation about the horizontal axis) by 0, 10, 20, 30, and 40 degrees and, separately, rotated about its own axis (pan) by the same angles; each source pose was paired with the 11 target poses of Table S1 and the displacement computed as above. The rim-estimating construction received the ground-truth mask; the frontal-assumption construction received the crop of the same mask, with $r_1/r_0 = 10$ as in the experiments, and for each target pose the one of its three viewpoint motions closest to the target camera, chosen from the normalized displacement between the two cameras and fixed before the runs. Table S2 and Fig. 3 give the result. Under tilt, the mean displacement of the rim-estimating construction stays between 4.1 and 5.3 px from 0 to 40 degrees, while that of the frontal-assumption construction rises from 26.8 to 79.6 px. Under pan both rise, from 4.1 to 53.0 px and from 26.8 to 73.0 px, because a rotation of the cylinder about its own axis leaves the rims and the vanishing point unchanged, and neither construction can recover it from them. The 26.8 px of the frontal-assumption construction at zero tilt is an offset that comes from the crop, canvas, and radius adaptation needed to give it a rendered image in place of a photograph, so the informative comparison is between the slopes of the two curves rather than their levels: the test shows what each construction does with a tilted input, not which is more accurate on a frontal one.

## IV. RETRIEVAL, ADAPTATION AND BASELINES

### A. BACKBONES AND FROZEN RETRIEVAL

Two backbones are used: DINO ViT-S/16 with the ImageNet self-supervised weights of [8], retained from the recipe of [4], and the vision tower of SigLIP 2-B (google/siglip2-base-patch16-224, 92.9 M parameters) [10]. SigLIP 2 images are corrected using exchangeable image file format (EXIF) metadata, resized with bicubic interpolation to 224 pixels on the longer side, letterboxed on black, and normalized with the official processor. Its pooled vectors are L2-normalized and ranked by cosine similarity. DINO uses the recovered Open Metric Learning (OML) preprocessing: longest-side resize to 224 pixels, white padding, and ImageNet normalization. Its unnormalized classification-token (CLS) features are ranked by Euclidean distance for both frozen and fine-tuned models. The development-only DINOv2-S comparison uses black letterboxing, L2-normalized CLS features, and cosine similarity. Retrieval is against one enrollment vector per class; the main benchmark contains 1,000 classes. No crop, fusion, or re-ranking is applied in the main results.

### B. ADAPTATION WITH SYNTHESIZED VIEWS

The synthesized views can be used to adapt the encoder in ways that update different fractions of its parameters, and Section VI-C compares the resulting recipes for the text-supervised backbone. Three recipes are used with SigLIP 2, ordered by how much of the encoder they change: full fine-tuning of the vision tower, at a fixed budget (T3) and at a validation-selected budget (G); LoRA adapters in the attention blocks; and a linear head over frozen features. DINO is fine-tuned with the recipe of [4] (R). Three seeds (42, 43, 44) are run for every configuration.

*Full fine-tuning of SigLIP 2 (T3 and G).* All parameters of the vision tower are trained from the pretrained initialization. Each update samples eight classes and two of their seven slots (the original plus six synthesized views), an effective batch of 16 without accumulation; classes are drawn without replacement within a cycle. Pooled vectors are L2-normalized, and for each anchor the loss is the softplus of the distance to the farthest positive in the batch minus the distance to the nearest negative, with squared Euclidean distance, no margin, and no memory bank. AdamW is used with a constant learning rate of $10^{-5}$ and weight decay 0.01, in single-precision floating-point (FP32) arithmetic with TensorFloat-32 (TF32) disabled and gradient checkpointing. No online augmentation is applied beyond the six synthesized slots.

The two budgets answer different questions. In the fixed-budget experiment (T3) the final checkpoint after 1,000 updates (100-class subset) or 4,000 updates (1,000 classes) is evaluated. These budgets were fixed in advance from calibration runs on a separate development set of 100 classes, in which no budget satisfied a joint criterion of near-maximal accuracy and flat training loss for all four pipelines, so the prespecified fallback budget was used; the budgets are matched across pipelines and are not a claim of convergence. In the validation-selected experiment (G) the same trajectory is checkpointed at 100, 250, 500, 1,000, 2,000, and 4,000 updates, and for each pipeline the checkpoint with the highest mean query-weighted top-1 over three seeds on the 1,384-class validation split is chosen, ties going to the earliest; only that checkpoint is evaluated on the benchmark. Benchmark queries are not used for any choice. The frozen model is the update-zero point of the same trajectory and is not one of the candidates.

*Frozen-feature and parameter-efficient adaptation (linear head and LoRA).* Two further recipes test whether the outcome of full fine-tuning is a property of the recipe rather than of the views. In the linear-head recipe the vision tower is frozen and a single 768 × 768 linear map without bias, initialized to the identity so that update zero reproduces the frozen model, is trained on the frozen features of the seven slots per class with the same sampler and loss as above, entirely on a central processing unit (CPU); AdamW with weight decay 0.01, the learning rate chosen from $\{10^{-4}, 10^{-3}\}$ and the checkpoint from {100, 250, 500, 1,000, 2,000, 4,000} updates by the validation rule of G. In the LoRA recipe, rank-8 adapters ($\alpha = 16$, no dropout) are added to the query, key, value, and output projections of the twelve self-attention blocks, 589,824 trainable parameters, with every other parameter frozen and verified unchanged; AdamW at $10^{-4}$ with weight decay 0.01, FP32, up to 1,000 updates with the checkpoint chosen on the validation split from {100, 250, 500, 1,000}, same sampler and loss. These recipes were added after the initial benchmark results had been inspected; their grids and validation-selection rules were fixed before evaluating the benchmark predictions for these recipes.

*DINO ViT-S/16 (R).* The recipe is that of [4] as implemented in the Open Metric Learning library: all parameters trained, unnormalized embeddings, soft-margin triplet loss with a memory-bank miner (50 batches, expansion 3), Adam at $10^{-5}$ without weight decay, a balanced sampler of 32 classes × 2 images, ImageNet normalization, and ranking by Euclidean distance. The same slots, queries, budgets (1,000 and 4,000 updates), seeds, and evaluation are used as for SigLIP 2. Because the optimizer, loss, sampler, and input normalization differ from the SigLIP recipe, differences between the two backbones cannot be attributed to the backbone alone; the DINO experiment reproduces the earlier recipe on public data, and pipelines are compared within a backbone.

### C. MULTI-VIEW ENROLLMENT

The views can also be used without any training: the six synthesized views of each enrollment image are embedded with the frozen SigLIP 2 and added to the gallery. Four ways of scoring a class are reported, all fixed in advance: the maximum cosine over the seven vectors (original plus six views), the cosine to the L2-normalized mean of the seven vectors, the cosine to a weighted mean in which each synthesized view is weighted by its cosine to the original photograph, and, as a diagnostic, the maximum over the six synthesized views alone. The 2D pipeline controls for the effect of simply having more vectors per class. The queries are unchanged, so the protocol remains one-shot with respect to the photographs supplied per class.

## V. BENCHMARK

### A. SOURCE DATA

WineSensed [12] is a public multimodal wine dataset; the source records list different noncommercial licenses, as noted in Data Availability. Its image component consists of user-uploaded label photographs from the Vivino platform: 996,808 distinct image files in the archive we verified by hash, each linked to a *vintage* identifier. The fixed metadata snapshot used here has 1,014,630 rows and no wine-name field, and its producer and year fields are populated only for the roughly 50,000 rows that belong to the sensory experiments of the original paper (96 distinct producer identifiers and 12 distinct years); for the remaining rows only the vintage identifier and the image link are available. The vintage identifier is therefore the only label available at scale, and the benchmark uses it as the class. Two consequences recur in the analysis: a vintage identifier can group photographs of different editions of a label (retailer-specific or redesigned labels of the same wine), and this snapshot does not support producer-level labels across the full benchmark.

The photographs are small (about 35 kB on average) and were taken by users under uncontrolled conditions; reflection, partial occlusion, and background clutter are common. The pose of the enrollment photographs is characterized below; that of the queries was not measured.

*Pose of the enrollment photographs.* Section III-E showed that the two cylindrical constructions diverge on tilted inputs, so whether the benchmark contains tilted enrollment photographs matters for interpreting their comparison. No pose annotation exists for these photographs, and the construction of Section III recovers neither the tilt nor the pan of the input as an angle. We therefore use an uncalibrated proxy that the SAM front end already computes: the ratio $\rho$ of the minor to the major axis of the ellipse fitted to each rim, averaged over the two rims of a source ($\bar{\rho}$). A rim seen edge-on projects to a line ($\rho = 0$), and $\rho$ grows with the elevation of the camera above or below the rim, approximately as its sine. Because the two rims of a label lie above and below the optical axis of a frontal photograph, $\bar{\rho}$ is not zero even at zero tilt: for a label 8–10 cm tall photographed from 25–40 cm, the rims lie 4–5 cm from the axis and project with $\rho$ of about 0.1–0.2 from perspective alone. Of the 1,000 sources, 783 have both rims fitted as ellipses, with $\bar{\rho}$ between 0.060 (5th percentile) and 0.208 (95th percentile) and a median of 0.130 (Fig. S1); for 215 sources at least one rim was too flat for an ellipse and was fitted as a line or a polynomial, and the front end failed on 2. These values are consistent with limited source tilt, but the proxy does not establish how many photographs are frontal. The tertiles of $\bar{\rho}$ over the 783 sources, at 0.098 and 0.160, were written to the experiment log before any stratified result was computed and are used in Sections VI-D and VI-G. The proxy is confounded with bottle shape, camera distance, and the visible fraction of the rim; it orders the sources and does not assign them angles.

### B. CLASSES AND SPLITS

Classes with at least two distinct images were eligible. Before the benchmark was sampled, every class that had been used in any earlier development experiment of this study was excluded, so that no benchmark class had been seen by any model selection. From the remaining classes, 1,000 were drawn with a fixed seed; for each class one image was drawn as the enrollment image and the others became queries, giving 4,295 queries. A nested subset of 100 classes with 351 queries is used for ablations. A separate split of 1,384 classes with 5,442 queries, drawn earlier, served for development, calibration, and checkpoint selection; it is not a blind test set and is reported only in that role. Exact duplicates were excluded by file hash and by decoded-pixel comparison among the 5,295 benchmark images. Near-duplicates between an enrollment image and its queries (photographs of the same bottle in the same session) cannot be excluded, and class identity was not verified by a human at the benchmark scale; Section VI-F reports what an audit of the errors found.

### C. METRICS AND STATISTICS

Top-1 is the primary outcome; top-5, top-10, and top-20 accuracy are also computed, both query-weighted and class-averaged (macro). The paper primarily presents query-weighted results and identifies macro results explicitly. Comparisons between two conditions use the paired per-query outcome: the number of queries rescued (wrong under the reference, correct under the condition) and broken, and the difference in accuracy with a 95% interval from a class-cluster bootstrap of 5,000 resamples, in which paired correctness differences are averaged over the three training seeds and whole classes are resampled. Query-weighted differences retain each sampled class's query count in the denominator; macro differences weight sampled classes equally. The interval therefore reflects the class composition of the benchmark, not the variation between training seeds, which is reported separately as the sample standard deviation (SD) over seeds. Frozen models are deterministic and are single runs. Intervals are not corrected for multiple comparisons; the original comparisons were specified before their formal runs. The whole-benchmark SAM-based versus frontal-assumption contrast added in Table VI during manuscript review is exploratory and uses retained predictions, with no new model run. The selection rules and thresholds of Section IV were fixed before benchmark evaluation of each respective arm, and the stratification of Section V-A before any stratified result was computed. Follow-up conditions added after the fixed-budget and validation-selected results had been seen included the flip-free 2D bank, the linear-head and LoRA recipes, the fourth bank of Section VI-D, and cosine-weighted enrollment. Each was given a selection rule that was fixed before it was evaluated on the benchmark. The 100-class subset is nested in the 1,000-class benchmark and is not an independent replicate.

### D. RELEASE

The released package (96 files, 11.7 MB) contains the identifier lists and SHA-256 hashes of the 12,121 distinct enrollment, query, subset, and validation images; the archive revision and metadata hash of the WineSensed release they were drawn from; the synthesis parameters of every slot of the three original pipelines and whether it fell back; the per-query top-20 rankings of 73 runs (the frozen model, the fixed-budget fine-tuning runs of both backbones on both splits, the validation-selected runs, and the multi-view enrollment runs of those pipelines); and scripts that rebuild the splits from the official release, verify the hashes, and recompute the corresponding tables from the rankings. The rankings of the arms added afterwards (the flip-free 2D control, the linear-head and LoRA recipes, the fourth bank of Section VI-D, and the cosine-weighted enrollment) are not in this snapshot and are available from the author. Images, features, and weights are not included. Before release the package was tested in a clean directory: the frozen result (4,069 of 4,295) and the three SAM-pipeline seeds of the fixed-budget experiment of Section VI-C (3,607, 3,707, and 3,704) were recomputed from the rankings and matched. The package and its versioned archive are publicly available at https://github.com/EdenHuang/winesensed-one-shot-benchmark.

## VI. RESULTS

### A. FROZEN BACKBONES

Without any training, the text-supervised backbone already solves most of the task and the self-supervised one does not. On the 1,000-class benchmark, SigLIP 2-B reaches 94.7% top-1 and 98.2% top-5, and its class-averaged top-1 of 94.80% shows that the query-weighted figure is not driven by a few large classes. DINO ViT-S/16 reaches 34.1% on the same benchmark, and DINOv2-S reaches 35.9% on the development split (Table I). Cropping the label with SAM before embedding helps the self-supervised model on the development split (35.9% to 47.3% for DINOv2-S) and does not change the text-supervised model (93.2% to 93.5%), so the SigLIP 2 results use the uncropped photograph throughout.

TABLE I
FROZEN BACKBONES, NO TRAINING

| Backbone | Input | Split | Top-1 | Top-5 | Top-10 | Top-20 |
|---|---|---|---|---|---|---|
| SigLIP 2-B | photo | benchmark (1,000 cl.) | 94.74 | 98.16 | 98.46 | 98.72 |
| DINO ViT-S/16 | photo | benchmark (1,000 cl.) | 34.09 | 46.87 | 52.36 | 58.53 |
| SigLIP 2-B | photo | development (1,384 cl.) | 93.22 | 97.26 | 98.16 | 98.77 |
| SigLIP 2-B | SAM crop | development (1,384 cl.) | 93.46 | 96.03 | 96.77 | 97.50 |
| DINOv2-S | photo | development (1,384 cl.) | 35.87 | 49.91 | 56.49 | 62.79 |
| DINOv2-S | SAM crop | development (1,384 cl.) | 47.32 | 61.45 | 66.67 | 71.04 |

Query-weighted accuracy (%).

### B. SELF-SUPERVISED BACKBONE: SYNTHESIS IS DECISIVE

For the self-supervised backbone, geometric view synthesis supplies most of the accuracy. The frozen DINO ViT-S/16 recognizes 34.1% of the queries. After 4,000 updates with the recipe of [4], 2D augmentation raises this to 44.3% (+10.2 points, 95% interval +8.2 to +12.4), with wide variation across seeds (38.4% to 49.5%). The three geometric pipelines raise it to 62.6–63.7%, a gain of 28.6–29.6 points over the frozen model and, for the SAM-based pipeline, 19.3 points over 2D augmentation (+17.0 to +21.6). Table II and Fig. 4(a) give the full results.

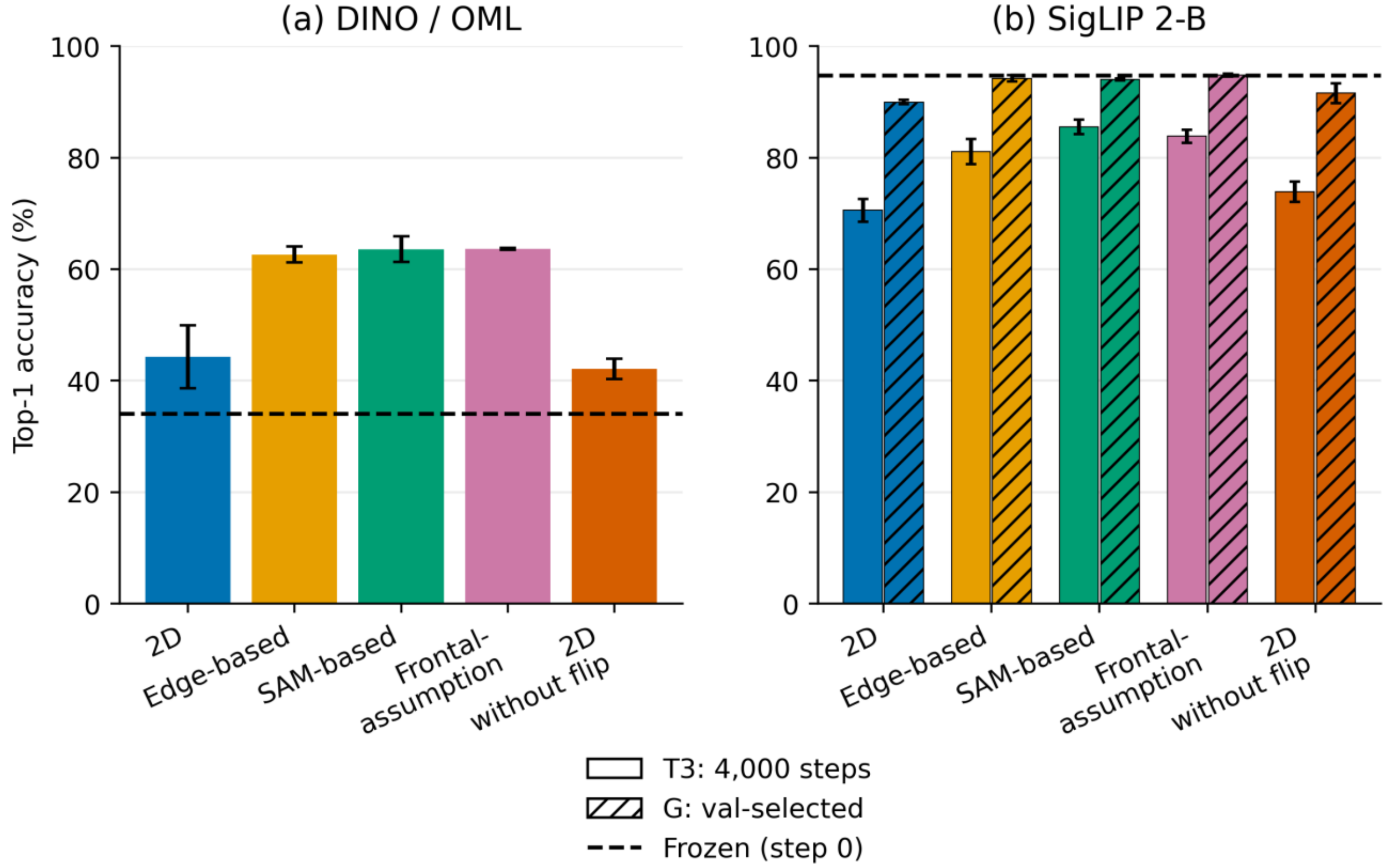

FIGURE 4. The two regimes. Top-1 accuracy on the 1,000-class benchmark after fine-tuning with each pipeline, mean and SD over three seeds; dashed lines are the frozen models. (a) DINO ViT-S/16 with the recipe of [4], 4,000 updates. (b) SigLIP 2-B after 4,000 updates (solid) and at the validation-selected checkpoint (hatched).

TABLE II
DINO VIT-S/16 FINE-TUNED WITH EACH PIPELINE, RECIPE OF [4]

| Pipeline | 1,000 cl., 4,000 upd. Top-1 | Top-5 | Δ vs frozen | 100 cl., 1,000 upd. Top-1 | Δ vs frozen |
|---|---|---|---|---|---|
| Frozen | 34.09 | 46.87 | — | 43.88 | — |
| 2D | 44.28 ± 5.62 | 56.24 | +10.19 [+8.15, +12.39] | 49.67 ± 2.14 | +5.79 [+0.10, +10.89] |
| 2D without flip | 42.11 ± 1.81 | 55.31 | +8.02 [+5.41, +10.51] | — | — |
| Edge-based [4] | 62.64 ± 1.46 | 75.10 | +28.55 [+25.82, +31.35] | 79.11 ± 1.19 | +35.23 [+24.95, +43.62] |
| SAM-based [5] | 63.58 ± 2.31 | 76.19 | +29.49 [+26.74, +32.31] | 83.00 ± 1.00 | +39.13 [+28.90, +47.57] |
| Frontal-assumption [7] | 63.66 ± 0.16 | 75.81 | +29.58 [+26.79, +32.43] | 80.25 ± 1.57 | +36.37 [+26.46, +44.61] |

Mean ± SD over three seeds; Δ against the frozen model with class-cluster 95% interval. Pairwise differences at 1,000 classes: SAM-based − 2D +19.3 [+17.0, +21.6]; SAM-based − edge-based +0.9 [−0.6, +2.6]; frontal-assumption − edge-based +1.0 [−0.9, +3.0]; 2D without flip − 2D −2.2 [−4.2, −0.5]; SAM-based − 2D without flip +21.5 [+18.8, +24.1]. On the 100-class subset, SAM-based − edge-based +3.9 [+0.6, +7.6].

Two further observations bear on later sections. First, the reported pairwise intervals do not resolve differences among the three geometric pipelines at this scale: the SAM-based and frontal-assumption pipelines are 0.9 and 1.0 points above the edge-based pipeline (−0.6 to +2.6 and −0.9 to +3.0), neither difference distinguishable from zero. On the 100-class subset after 1,000 updates the SAM-based pipeline does separate from the edge-based one (83.0% against 79.1%, a difference of 3.9 points, +0.6 to +7.6); Section VI-D examines what the two front ends contribute. Second, the horizontal flip in the 2D control is not what holds it back on this backbone: removing the flip lowers accuracy to 42.1%, 2.2 points below the original control (−4.2 to −0.5) and 21.5 points below the SAM-based pipeline (−24.1 to −18.8).

In this regime, then, the finding of [4] reproduces on public data, and so does the direction of the finding of [7], that a cylinder model beats 2D augmentation.

### C. TEXT-SUPERVISED BACKBONE: THE GAIN DEPENDS ON THE ADAPTATION RECIPE

For the text-supervised backbone the same views produce a different outcome, with substantial differences across the tested adaptation recipes. We report the recipes in the order they were run, from full fine-tuning to a head over frozen features, and then bring them together.

*Fixed budget.* Full fine-tuning for the fixed budget of 4,000 updates loses accuracy with every pipeline (Table III, Fig. 4(b)): 2D augmentation falls from the frozen 94.7% to 70.6% (−24.2 points, 95% interval −27.5 to −21.1), and the SAM-based pipeline, the best of the four, to 85.5% (−9.2, −12.3 to −6.6). The ordering among the pipelines is the one seen with DINO: the SAM-based pipeline is 4.4 points above the edge-based one (+1.9 to +7.1) and 15.0 above 2D. The flip-free 2D control retains 3.3 points more than the original control (+1.6 to +5.0) but remains 11.6 points below the SAM-based pipeline (−14.5 to −8.7). On the 100-class subset after 1,000 updates, the SAM-based and frontal-assumption pipelines are within noise of the frozen model (97.0% and 97.2% against 97.4%; SAM-based: −0.5, −2.1 to +0.8) while 2D loses 10.4 points. The subset has ten times fewer classes for a quarter of the updates, so each class is seen about 2.5 times more often than on the full benchmark, so class count and exposure per class change together. This comparison does not isolate why the full benchmark loses more accuracy, and neither budget establishes convergence.

TABLE III
SIGLIP 2-B FINE-TUNED WITH EACH PIPELINE, FIXED BUDGET

| Pipeline | Replaced slots | 1,000 cl., 4,000 upd. Top-1 | Top-5 | Δ vs frozen | 100 cl., 1,000 upd. Top-1 | Δ vs frozen |
|---|---|---|---|---|---|---|
| Frozen | — | 94.74 | 98.16 | — | 97.44 | — |
| 2D | 0 | 70.56 ± 2.06 | 79.08 | −24.18 [−27.46, −21.11] | 86.99 ± 2.02 | −10.45 [−17.43, −5.61] |
| 2D without flip | 0 | 73.89 ± 1.84 | 81.67 | −20.85 [−24.10, −17.83] | — | — |
| Edge-based [4] | 3,412 | 81.10 ± 2.26 | 87.83 | −13.64 [−17.24, −10.50] | 91.74 ± 1.78 | −5.70 [−10.15, −2.40] |
| SAM-based [5] | 42 | 85.51 ± 1.33 | 92.61 | −9.23 [−12.34, −6.63] | 96.96 ± 0.33 | −0.48 [−2.11, +0.83] |
| Frontal-assumption [7] | 12 | 83.84 ± 1.16 | 91.20 | −10.90 [−14.06, −8.19] | 97.25 ± 0.33 | −0.19 [−1.39, +0.89] |

Mean ± SD over three seeds; Δ against frozen with class-cluster 95% interval; replaced slots out of the 6,000 synthesis slots. Pairwise differences at 1,000 classes: SAM-based − edge-based +4.4 [+1.9, +7.1]; 2D without flip − 2D +3.3 [+1.6, +5.0]; SAM-based − 2D without flip +11.6 [+8.7, +14.5].

*Validation-selected budget.* Choosing the checkpoint on the validation split instead of fixing it removes the loss for the geometric pipelines but does not turn it into a gain. On the development split, accuracy is below the frozen 93.2% at every checkpoint for every pipeline, including the first at 100 updates (92.6–93.0% for the geometric pipelines, 86.9% for 2D), and declines from there, slowly for the geometric pipelines and quickly for 2D (Fig. 5). The rule therefore selected 100 updates for 2D, the edge-based, and the frontal-assumption pipelines and 250 for the SAM-based pipeline. Evaluated once on the benchmark at those checkpoints (Table IV), the three geometric pipelines are within 0.7 points of the frozen model, with intervals that include zero, while 2D remains 4.8 points behind (−6.7 to −3.2) and the flip-free control 3.2 behind (−4.9 to −1.8), 1.6 above the original control and 2.5 below the SAM-based pipeline. Parity does not mean the adapted models are the same model as the frozen one: the SAM-based pipeline at 250 updates rescues 81–84 queries per seed and breaks 106–123.

Fig. 5 also plots the mean cosine between each validation image's adapted and frozen embeddings. The embeddings move away from the frozen coordinates within the first 100 updates (mean cosine 0.35 for 2D, 0.40–0.42 for the geometric pipelines) and keep moving (0.13–0.17 at 4,000 updates), and accuracy declines in step. At the first checkpoint, however, the four pipelines have moved by similar amounts and differ by six points in accuracy, so the distance from the frozen coordinates is not by itself what determines the loss. We report the drift as a description of the trajectories, not as a mechanism.

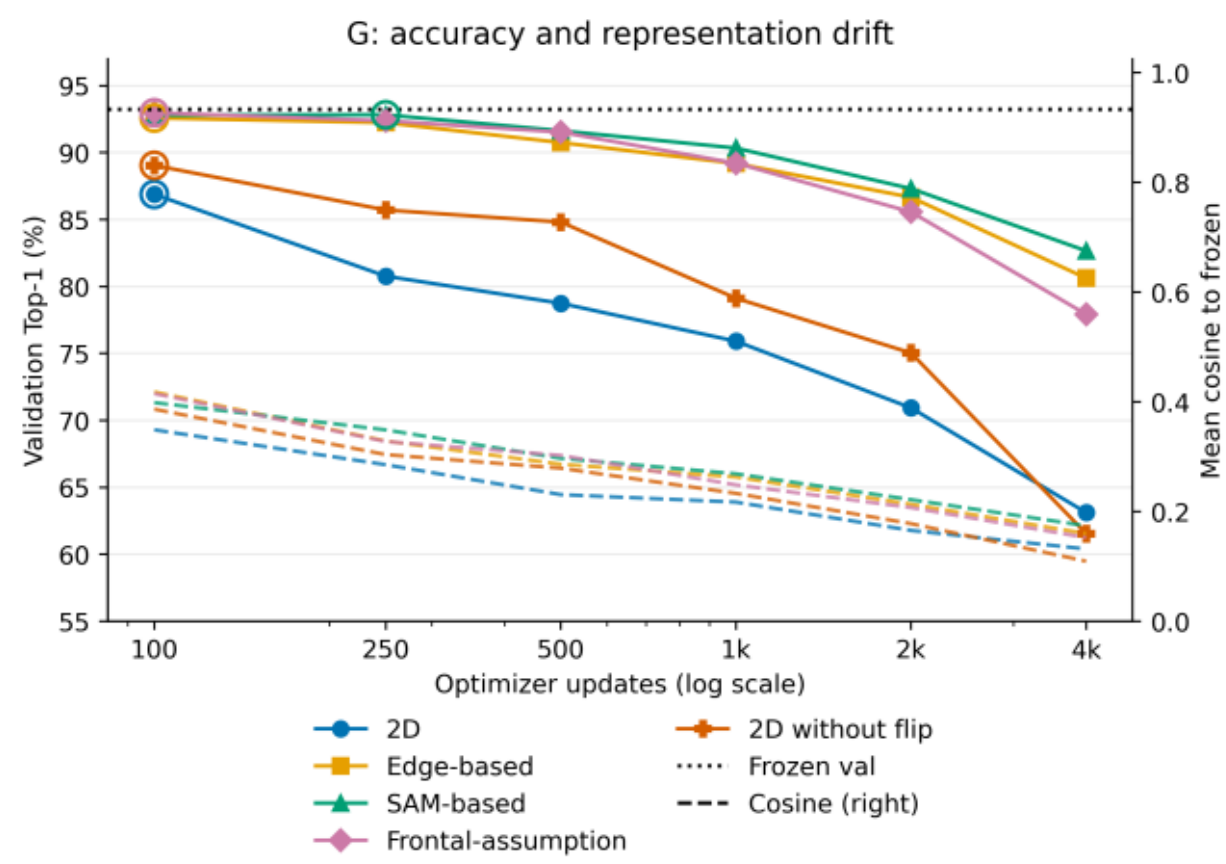

**FIGURE 5.** Validation accuracy and embedding drift during SigLIP 2-B fine-tuning, averaged over three seeds. Solid lines: top-1 on the 5,442 validation queries (left axis), with the frozen model as the dotted line; open circles mark the checkpoint chosen for each pipeline. Dashed lines: mean cosine between the adapted and the frozen embedding over the 6,826 validation images (right axis).

TABLE IV
SIGLIP 2-B, VALIDATION-SELECTED CHECKPOINT, 1,000 CLASSES

| Pipeline | Selected updates | Top-1 | Rescued / broken (seeds 42; 43; 44) | Δ vs frozen |
|---|---|---|---|---|
| Frozen | 0 | 94.74 | — | — |
| 2D | 100 | 89.97 ± 0.40 | 50/274; 50/241; 44/244 | −4.77 [−6.67, −3.15] |
| 2D without flip | 100 | 91.56 ± 1.74 | 54/172; 53/272; 66/139 | −3.18 [−4.94, −1.84] |
| Edge-based [4] | 100 | 94.24 ± 0.59 | 73/122; 83/83; 83/98 | −0.50 [−1.94, +0.70] |
| SAM-based [5] | 250 | 94.03 ± 0.18 | 81/108; 84/123; 81/106 | −0.71 [−2.12, +0.52] |
| Frontal-assumption [7] | 100 | 94.80 ± 0.27 | 84/68; 78/83; 81/84 | +0.06 [−1.01, +1.10] |

Top-1 (%), mean ± SD over three seeds; Δ against frozen with class-cluster 95% interval. Pairwise differences: 2D without flip − 2D +1.6 [+0.8, +2.5]; SAM-based − 2D without flip +2.5 [+1.6, +3.5].

*Multi-view enrollment.* Using the views without training yields only small gains under maximum aggregation. Adding the six synthesized views of each enrollment image to the gallery of the frozen model and scoring each class by the maximum cosine over its seven vectors changes top-1 by +0.05 points for the SAM-based pipeline and +0.02 for the frontal-assumption pipeline, with two and one queries rescued and none broken (Table S3, Fig. 6). The 2D control changes it by +0.23 (+0.07 to +0.42; 12 rescued, 2 broken), the only interval that excludes zero, and its flip-free variant by +0.12 (−0.05 to +0.28), but these intervals alone do not establish a difference between the two 2D controls. Averaging the seven vectors lowers accuracy for every pipeline (−0.8 to −4.9 points), and so does the cosine-weighted mean, except for the flip-free control, whose interval includes zero. The diagnostic condition is the informative one: when the six synthesized views alone replace the photograph as the enrollment, accuracy drops by 22.0 points for the SAM-based pipeline and 14.4 for the frontal-assumption pipeline, but by only 1.6 for 2D. In the embedding space of SigLIP 2, a geometrically synthesized view of a label is a poor stand-in for a photograph of it, while a 2D-augmented copy is not; Section VI-G measures this directly.

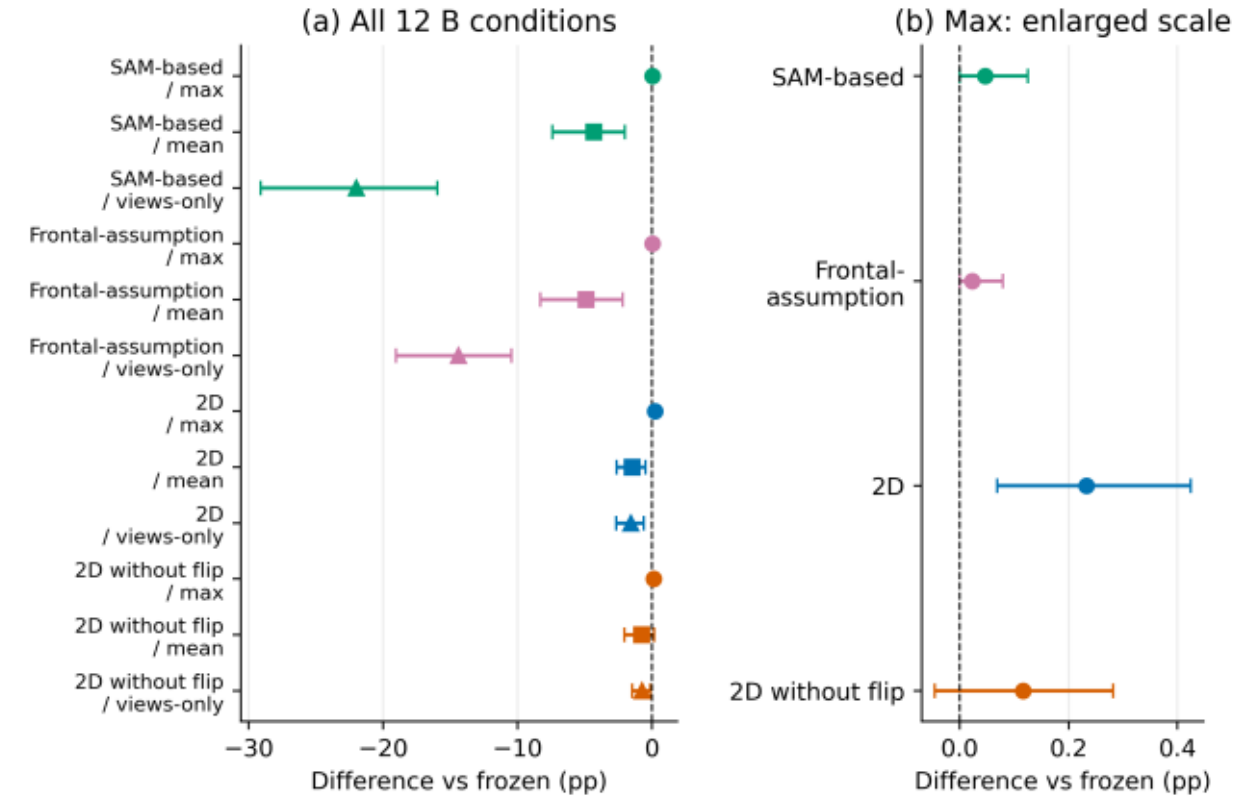

**FIGURE 6.** Multi-view enrollment with the frozen SigLIP 2-B model. Change in top-1 against the single-photograph gallery for each pipeline and aggregation rule, with class-cluster 95% intervals; (b) enlarges the maximum-aggregation points of (a).

*Adaptation that preserves the encoder.* The two recipes of Section IV-B that leave the encoder untouched or nearly so show that the views do carry something the frozen embedding does not use (Table V). With the vision tower frozen and only a linear projection trained on the synthesized views, every pipeline ends above the frozen model, and for the two SAM-localized geometric pipelines the gain is established: 96.0% for the SAM-based pipeline (+1.24 points, +0.52 to +1.96; 69–74 queries rescued and 17–19 broken per seed) and 96.0% for the frontal-assumption pipeline (+1.30, +0.58 to +1.97). The 2D control, its flip-free variant, and the edge-based pipeline gain 0.3–0.6 points with intervals that include zero; the SAM-based pipeline is 0.92 points above 2D (+0.40 to +1.50) and indistinguishable from the frontal-assumption pipeline (−0.05, −0.21 to +0.10). The validation rule chose the learning rate $10^{-4}$ for every pipeline and the full 4,000 updates for the three geometric pipelines, against 100 for 2D and 500 for the flip-free control; on the validation split the geometric heads reach 94.3–94.4% against the frozen 93.2%. With LoRA adapters, which do update the encoder, every interval includes zero, the largest gain being +0.73 (−0.20 to +1.67) for the frontal-assumption pipeline; the checkpoints were chosen at 100–500 updates, and per seed about 80 queries are rescued against 54–117 broken, where the linear head broke 10–26.

TABLE V
ADAPTATION OF SIGLIP 2-B WITH THE ENCODER PRESERVED, 1,000 CLASSES

| Recipe | Pipeline | Selected updates | Top-1 | Rescued / broken (seeds 42; 43; 44) | Δ vs frozen |
|---|---|---|---|---|---|
| Frozen | — | 0 | 94.74 | — | — |
| Linear head (lr $10^{-4}$) | 2D | 100 | 95.06 ± 0.08 | 24/14; 25/10; 29/12 | +0.33 [−0.12, +0.76] |
| Linear head (lr $10^{-4}$) | 2D without flip | 500 | 95.34 ± 0.09 | 43/18; 45/23; 45/15 | +0.60 [−0.01, +1.19] |
| Linear head (lr $10^{-4}$) | Edge-based [4] | 4,000 | 95.34 ± 0.13 | 46/26; 50/19; 48/22 | +0.60 [−0.07, +1.16] |
| Linear head (lr $10^{-4}$) | SAM-based [5] | 4,000 | 95.98 ± 0.05 | 71/19; 74/18; 69/17 | +1.24 [+0.52, +1.96] |
| Linear head (lr $10^{-4}$) | Frontal-assumption [7] | 4,000 | 96.03 ± 0.12 | 67/16; 73/18; 77/16 | +1.30 [+0.58, +1.97] |
| LoRA r = 8 | 2D | 100 | 94.55 ± 0.13 | 78/91; 83/92; 81/83 | −0.19 [−1.81, +1.10] |
| LoRA r = 8 | 2D without flip | 250 | 93.99 ± 0.31 | 73/117; 74/108; 75/93 | −0.75 [−2.41, +0.57] |
| LoRA r = 8 | Edge-based [4] | 250 | 94.35 ± 0.17 | 73/98; 77/90; 74/86 | −0.39 [−2.07, +0.92] |
| LoRA r = 8 | SAM-based [5] | 500 | 94.98 ± 0.15 | 88/75; 85/82; 84/69 | +0.24 [−0.93, +1.32] |
| LoRA r = 8 | Frontal-assumption [7] | 250 | 95.47 ± 0.09 | 91/58; 82/55; 88/54 | +0.73 [−0.20, +1.67] |

Top-1 (%), mean ± SD over three seeds; learning rate and checkpoint chosen on the validation split; Δ against frozen with class-cluster 95% interval. Under the linear head, SAM-based − 2D is +0.92 [+0.40, +1.50] and SAM-based − frontal-assumption −0.05 [−0.21, +0.10]; under LoRA, +0.43 [−0.18, +1.19] and −0.49 [−1.11, +0.01].

Taken together, the results differ substantially across adaptation recipes. A linear head over frozen features gains 1.2–1.3 points from the two SAM-localized geometric pipelines and less than a point, with intervals that include zero, from the edge-based pipeline and from 2D views; LoRA adapters show no gain that the intervals support; full fine-tuning stopped on the validation split returns to the frozen accuracy; full fine-tuning for a fixed budget loses 9 to 24 points; and the views add at most a quarter of a point as enrollment. The synthesized views therefore carry information that the frozen embedding does not use, but it is worth about one point, and among the tested recipes its clearest gain comes from a linear head over frozen features. What the choice of pipeline controls is the size of the gain in the first case and the size of the loss in the last: after 4,000 updates of full fine-tuning the SAM-based pipeline retains 85.5 points where the flip-free 2D control retains 73.9.

### D. FRONT-END RELIABILITY

The SAM-based front end synthesizes all six views for 993 of the 1,000 sources; the edge-based front end does so for 431. This section asks what that difference in coverage buys, and whether anything else separates the geometric pipelines.

*Coverage, or geometry?* Table S4 splits each pipeline's queries by whether their source was fully synthesized. For the edge-based pipeline the two halves are large enough to compare, and the trained model scores the same on both: after 4,000 SigLIP updates 81.1% on the fully native sources (81.12 ± 1.23) and 81.1% on the fallback sources (81.08 ± 3.74), and after 4,000 DINO updates 62.1% and 63.1%. The two halves are not equally hard, however: the frozen SigLIP 2 model scores 96.0% on the queries of the fully native sources and 93.6% on the rest, against 94.7% overall, so relative to the frozen model the edge-based pipeline loses slightly more on the sources where its own localization succeeded (−14.9 points) than on those where it failed (−12.5). What Table S4 establishes is that the loss of the edge-based pipeline is not confined to the sources on which its localization failed. Whether the SAM front end also yields better training views where both front ends succeed is a separate question, and a paired comparison on the 2,043 queries of the 431 sources that the edge-based front end synthesized natively does not settle it. Under fixed-budget full fine-tuning the SAM-trained model is 5.2 points ahead on those queries (+1.5 to +9.1), but it is also 3.7 points ahead (+0.3 to +7.3) on the queries of the other 569 sources, whose slots the edge-based bank fills with 2D images, so at this budget the difference between the two banks acts on every query through the shared encoder and cannot be attributed to the views of the sources where both succeeded. At the validation-selected checkpoint the two models are indistinguishable on the native sources (−0.6 points, −1.8 to +0.3), and so are the two DINO models (+1.3, −0.9 to +3.9). An earlier development study on 100 sources agrees: the SAM front end supplied all six views for 98 sources and the edge-based one for 41. Native-slot coverage was 588 of 600 versus 249 of 600, a difference of 56.5 points (95% interval 47 to 66), and on the 41 sources where both succeeded the recognition rates of the two pipelines could not be distinguished, with an interval too wide to exclude a difference of several points. The coverage contribution of the SAM front end is therefore established, since on real user photographs an edge detector designed for cleaner images fails more than half the time; a further contribution through better views where both front ends succeed is neither established nor excluded by this benchmark. The fallback subsets of the SAM-based (13 queries) and frontal-assumption (3 queries) pipelines are too small to interpret. As a check on the masks themselves, the SAM mask has a mean intersection over union (IoU) of 0.927 on 50 of 200 photographs for which the author annotated the label region (48 above 0.5, 45 above 0.75); the remaining 150 were not annotated, and the figure is reported as a sample only.

*Crossed comparison of front end and construction.* The SAM-based and frontal-assumption pipelines share the SAM front end but differ in construction and rendering conventions; the SAM-based and edge-based pipelines share the construction and differ only in the front end. A fourth bank completes the 2 × 2: the frontal-assumption construction fed with the label region found by the edge-based front end (the bounding box of the region enclosed by the two fitted rims and the tangent edges, with pixels outside the region set to black), under the same fallback rule. It falls back on 3,086 of 6,000 slots against 3,412 for the edge-based pipeline, which we attribute to the construction needing only the region and not the vanishing point.

Table VI gives the results after 4,000 updates. With SigLIP 2, changing the front end from edge-based to SAM raises accuracy by 4.4 points under the rim-estimating construction (+1.9 to +7.1) and by 2.5 under the frontal-assumption construction (+0.3 to +4.8), whereas changing the construction does less: with the edge-based front end the frontal-assumption construction is 0.2 points above the rim-estimating one (−1.3 to +2.0), and with the SAM front end 1.7 points below it. A whole-benchmark paired contrast added during manuscript review gives SAM-based minus frontal-assumption +1.67 points (+0.13 to +3.16). The wider within-stratum intervals in Table S5 do not negate this aggregate difference. With DINO the four cells lie within two points of one another (62.6% and 61.7% with the edge-based front end, 63.6% and 63.7% with SAM), and neither the front-end effect (+0.9 and +2.0 points) nor the construction effect (−0.9 and +0.1) is distinguishable from zero. Thus, front-end choice affects coverage and fixed-budget SigLIP 2 accuracy, while the two constructions also differ under that recipe. These are pipeline comparisons: the crop, canvas, and scale differences prevent attribution to cylinder geometry alone.

TABLE VI
FRONT END × CONSTRUCTION AFTER 4,000 UPDATES, 1,000 CLASSES

| Backbone, recipe | Front end | Rim-estimating construction [4] | Frontal-assumption construction [7] |
|---|---|---|---|
| SigLIP 2-B, T3 | Edge-based | 81.10 ± 2.26 (2,588) | 81.33 ± 1.29 (2,914) |
| SigLIP 2-B, T3 | SAM | 85.51 ± 1.33 (5,958) | 83.84 ± 1.16 (5,988) |
| DINO ViT-S/16, R | Edge-based | 62.64 ± 1.46 (2,588) | 61.71 ± 0.11 (2,914) |
| DINO ViT-S/16, R | SAM | 63.58 ± 2.31 (5,958) | 63.66 ± 0.16 (5,988) |

Top-1 (%), mean ± SD over three seeds; native slots out of 6,000 in parentheses (identical for both backbones). Paired differences with class-cluster 95% intervals, SigLIP 2: SAM − edge-based +4.41 [+1.92, +7.08] with the rim-estimating and +2.52 [+0.26, +4.77] with the frontal-assumption construction; frontal-assumption − rim-estimating +0.23 [−1.32, +1.97] with the edge-based front end. An exploratory whole-benchmark contrast computed during manuscript review is rim-estimating − frontal-assumption with SAM localization: +1.67 [+0.13, +3.16] for SigLIP 2. DINO: SAM − edge-based +0.94 [−0.6, +2.6] and +1.96 [−0.05, +4.12]; frontal-assumption − rim-estimating −0.93 [−2.14, +0.25] with the edge-based front end. The DINO runs of the edge-based × frontal-assumption cell use a numerically stable evaluation of the soft-margin loss term $\log(1 + e^x)$, after the original evaluation failed at updates 200, 105, and 150 for seeds 42, 43, and 44, respectively. The stable form matches the loss and gradients through the 199 valid updates of the seed-42 equivalence check; the remaining recipe is unchanged.

*Stratified by the pose proxy.* To test whether the construction contrast varies with the source-pose proxy, Table S5 and Fig. S2 split the comparison between the two constructions that share the SAM front end by the rim-ratio tertiles of Section V-A. Within each tertile and under all three recipes the paired intervals include zero, and the prespecified test, whether their difference grows from the lowest to the highest tertile, has an interval that includes zero for every recipe: +0.1 points (−4.6 to +4.8) for SigLIP 2 at 4,000 updates, −0.4 (−2.9 to +1.8) at the validation-selected checkpoint, and +2.6 (−0.3 to +5.3) for DINO, where the sign is the one Table S2 would predict but the interval is not. The tertiles are not equally easy, since the frozen SigLIP 2 model scores 96.1%, 95.3%, and 91.6% on them and frozen DINO 37.8%, 36.7%, and 26.3%, so the ratio orders the sources by something that matters to every model; over the observed proxy range, however, the analysis does not establish a change in the difference between constructions. The same holds with the edge-based front end (+0.6 points, −3.9 to +5.4, for SigLIP 2 at 4,000 updates).

### E. RE-RANKING AND OCR

Two second-stage methods common in product retrieval were tried as development-stage controls, and neither provides a clear accuracy gain in these controls. In local-feature re-ranking, the top-K candidates of the frozen SigLIP 2 are matched to the query with SIFT [23], LoFTR [24], or SuperPoint with LightGlue [25], [26], and the candidate with the most matches, or most RANSAC inliers, is promoted. In a development study of 100 queries and 2,968 pairs, direct promotion gave 93–94 correct with SIFT, 90–93 with LoFTR, and 93–95 with SuperPoint+LightGlue, against 94 without re-ranking; the best setting rescued two queries and broke one. A conservative rule was therefore fixed on a separate calibration set of 300 queries and applied once to a held-out set of 300: with SuperPoint+LightGlue at $K = 5$, the candidate with the most raw matches is promoted only if the embedding margin between the original first and second candidates is at most $t = 0.04$, so that the embedding itself was uncertain, and the relative advantage of the new candidate in match count, (new − original)/new, is at least $u = 0.50$, the thresholds chosen from a grid of four values of t and three of u plus the always-keep rule. On the held-out set it raised 283 correct to 285 (three rescued, one broken, one wrong answer exchanged for another), a difference of +0.7 points with a 95% interval of −0.67 to +2.00, and by the stopping rule fixed in advance no further variants were tried. The nine queries that re-ranking broke in development show why: the wrongly promoted candidate shared most of its layout with the query, typically the same producer's template with a different product name or year, so match counts favor the shared template over the small region that differs.

The OCR baseline uses EasyOCR 1.7.2 (CRAFT detector, Latin recognizer, six European languages) on grayscale images reduced to 640 pixels, keeps text boxes with confidence at least 0.5, normalizes the strings to lower-case alphanumeric tokens after Unicode normalization and accent removal, and scores each candidate by the cosine similarity of character-trigram TF-IDF vectors, with document frequencies computed on the enrollment images only; candidates within the embedding's top-5, 10, or 20 are re-ordered by this score. It reached 88 correct at K = 5 and 84 at K = 10 and 20 on the same 100 queries, against 94 for the embedding alone. These are single development studies of particular implementations, and neither method enters the main results.

### F. ERROR ANALYSIS

The frozen SigLIP 2 model misses 226 of 4,295 queries, and their true classes are usually close: 107 (47%) are ranked second, 147 (65%) within the top five, and the median rank is 3, although 36 (16%) are ranked beyond 100 (Fig. 7(a)).

An obvious hypothesis is that the errors concentrate on classes whose enrollment images are visually near-duplicates of another class's. It does not hold. A threshold fixed in advance at the 99th percentile of the cosine similarity between enrollment images of different classes on the development split (0.767, from 957,036 pairs) flags 129 of the 226 errors (57%), but applied to the second-ranked enrollment image of the 4,069 correct queries it flags 2,280 (56%), and at the 95th and 99.9th percentiles the errors are flagged less often than the correct queries. Enrollment images that are near-duplicates of another class's in the embedding are common throughout the benchmark, and the model resolves most of them.

What does explain the errors is visible only on inspection. Fifty errors were drawn with a fixed seed and inspected by one annotator (the author) with the query, its enrollment image, and the top-ranked enrollment image side by side (Table VII). Eleven are confusions between labels of the same layout and sixteen are errors in which the top-ranked label has a different layout from the query; these are the model's errors. Twenty-one have a visibly different label from their assigned enrollment image, consistent with an edition change or a possible metadata mismatch; visual inspection alone does not verify the underlying wine identity. Two are unreadable. Neither category establishes that retrieval is impossible. Nine of the twenty-one belong to one class whose enrollment image is a supermarket-branded Cava while its queries show the producer's own labels; these nine were flagged by a multimodal language model and individually confirmed by the author during the final manuscript review. That class has 45 queries, of which 31 are errors, 13.7% of all errors on the benchmark. Excluding it leaves 195 errors with the same rank profile (86 at rank 2, 36 beyond 100, median 3) and a near-duplicate rate of 51%; the class-cluster bootstrap used for accuracy comparisons preserves within-class dependence. The audit counts remain descriptive and should not be extrapolated to an irreducible error rate.

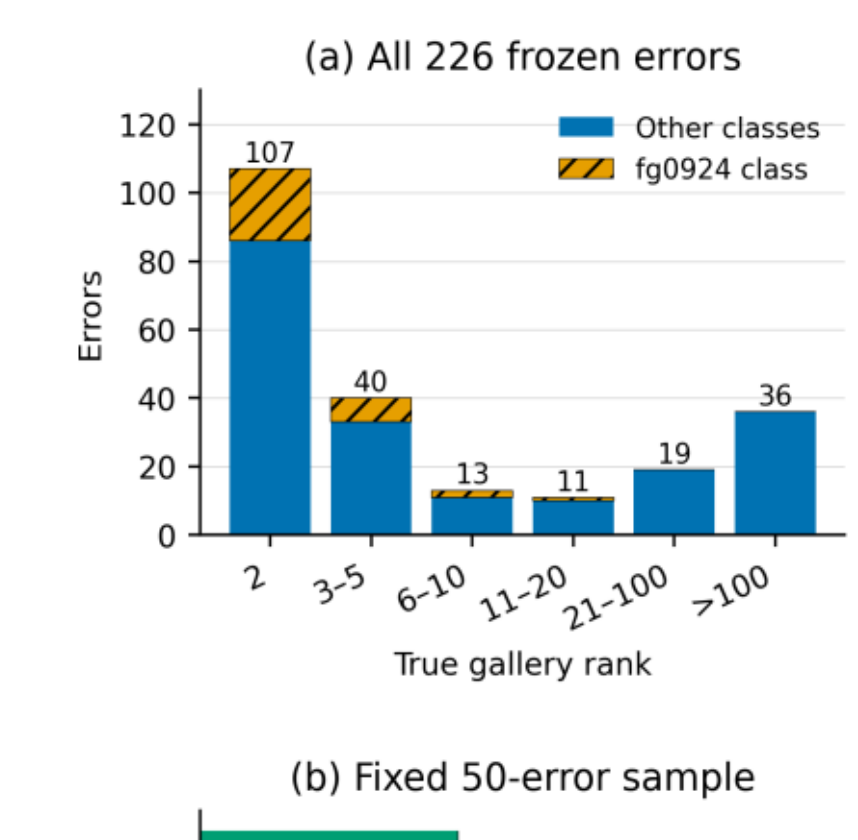

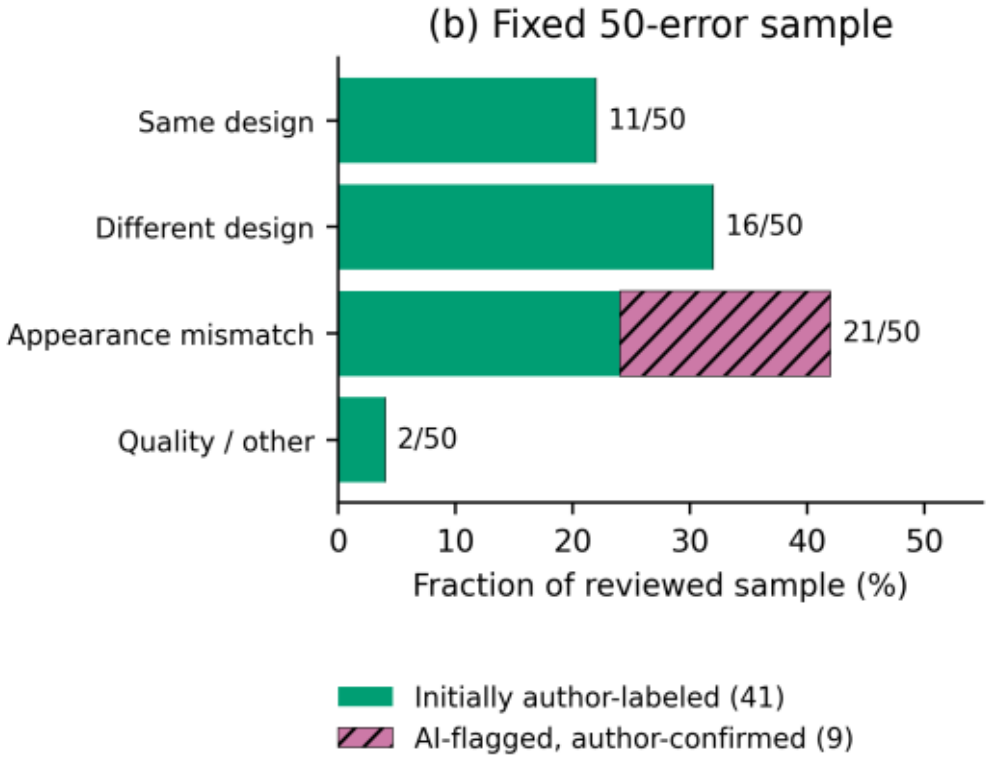

FIGURE 7. The 226 residual errors of the frozen SigLIP 2-B model. (a) Rank of the true class; the 31 errors of class fg0924 are hatched. (b) Categories assigned in the audit of 50 errors; the nine hatched cases are the query–enrollment appearance mismatches of one class that were flagged by a multimodal language model and confirmed by the annotator (Section VI-F). AI denotes artificial intelligence.

TABLE VII
AUDIT OF 50 RESIDUAL ERRORS OF THE FROZEN MODEL

| Category | n | Share | True class at rank 2 | Flagged near-duplicate |
|---|---|---|---|---|
| Same layout, sibling label confused | 11 | 22% | 8 | 10 |
| Different layout, model error | 16 | 32% | 6 | 6 |
| Query–enrollment appearance mismatch | 21 | 42% | 8 | 15 |
| Unreadable query / other | 2 | 4% | 0 | 0 |

The audit identifies query–enrollment appearance mismatch as a recurring issue, but it does not estimate an accuracy ceiling: identity was not independently verified, and visually different labels need not be impossible to match. The near-duplicate flag hits 10 of the 11 sibling confusions but also 15 of the 21 appearance mismatches, so this embedding-based flag alone does not distinguish those categories. The 50-case counts are descriptive, with nine confirmed mismatches concentrated in one class.

### G. DISTANCE BETWEEN SYNTHESIZED VIEWS AND PHOTOGRAPHS

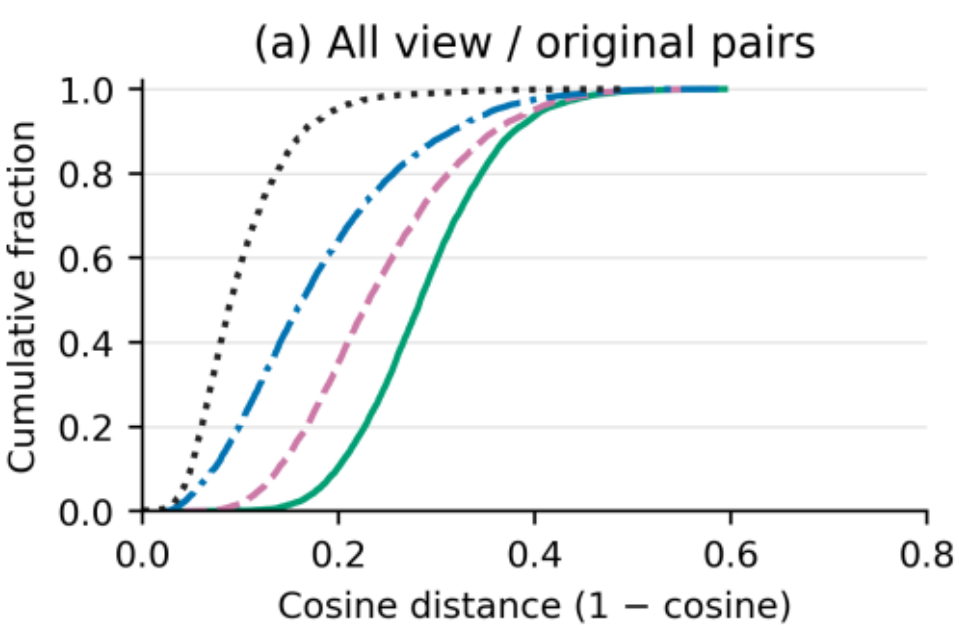

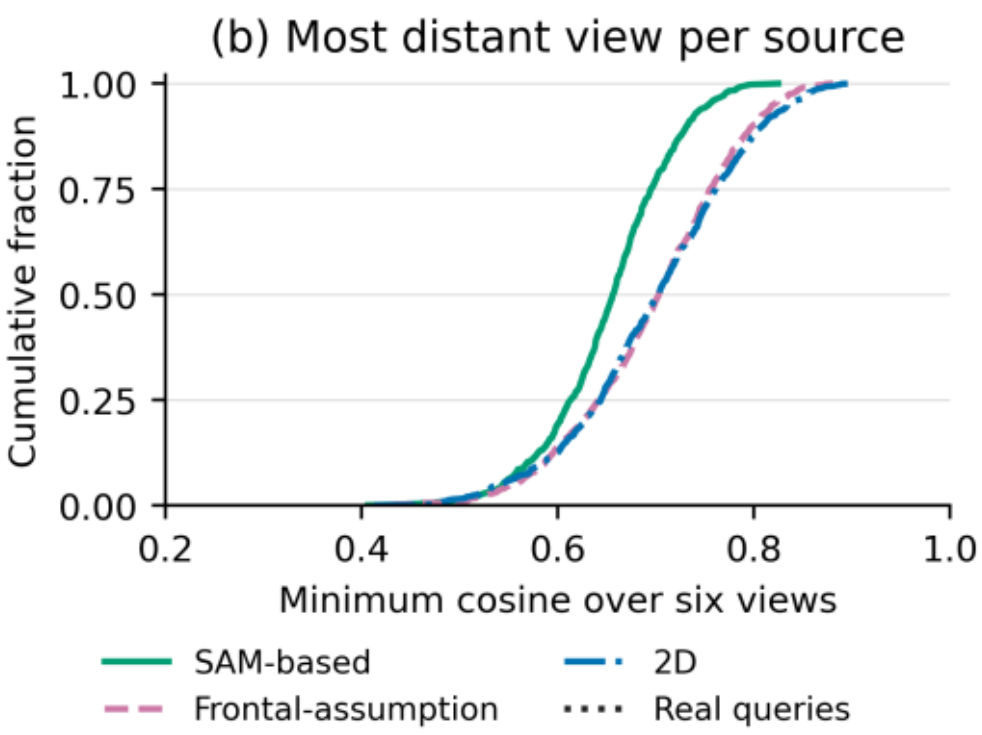

**FIGURE 8. Distance between synthesized views and photographs in the frozen SigLIP 2-B space. (a) Cumulative distribution of the cosine distance from each enrollment photograph to its six synthesized views, for each pipeline, and from each real query to the enrollment photograph of its class. (b) Minimum cosine over the six views of each source.**

The enrollment result of Section VI-C suggested that the frozen encoder does not treat a synthesized view as a photograph of the same label, and the feature space confirms it. Table S6 and Fig. 8 measure, in the frozen SigLIP 2 space, how far each synthesized view lies from the photograph it was made from, against how far a real query photograph lies from the enrollment photograph with the same dataset vintage identifier. The real photographs lie closest, with a mean cosine distance of 0.10 between a query and its enrollment image, followed by the 2D copies at 0.18 and the flip-free 2D copies at 0.15; the two geometric pipelines lie farthest, at 0.24 for the frontal-assumption pipeline and 0.29 for the SAM-based pipeline, and their 5th percentiles (0.12 and 0.18) lie near the 95th percentile of the real pairs (0.20). Taking for each source only its most distant view, the minimum cosine over six views averages 0.65 for the SAM-based pipeline against 0.70 for the other two.

The encoder thus sees a geometrically re-projected label, composited on a new background, as a larger change than the observed query–enrollment differences within a dataset class. This is a statement about the frozen feature space, not about the training dynamics, but it is consistent with the enrollment result of Table S3 and with the direction of the adaptation results: the images that the geometric pipelines add to training are, to this encoder, unlike the photographs it will be asked to match. No frozen DINO features of the synthesized views were available, so the measurement is reported for SigLIP 2 only.

Split by the rim-ratio tertiles of Section V-A (per-source means, sources weighted equally), the distance of the SAM-based views from their source is the same in all three tertiles (0.288, 0.285, and 0.281), whereas that of the frontal-assumption views is smallest in the lowest rim-ratio tertile (0.229) and larger in the other two (0.254 and 0.246). The difference between the two pipelines falls from 0.059 (0.052 to 0.067) in the lowest tertile to 0.035 (0.028 to 0.042) in the highest, a change of −0.024 (−0.035 to −0.015) under independent resampling of the two strata. This association between the proxy and feature distance differs between the pipelines. It is qualitatively consistent with the rendered-cylinder result, but the uncalibrated proxy and rendering differences preclude a causal interpretation. No corresponding proxy-dependent trend in the recognition contrast is established (Table S5).

### H. COST

On the RTX 3080 used for all experiments, embedding one query with SigLIP 2-B takes a median of 12.0 ms including file decoding and preprocessing (95th percentile 15.8 ms) at batch size 1 in FP32, and with DINO ViT-S/16 9.3 ms (9.9 ms); exact search over the 1,000 enrollment vectors with a single-threaded CPU FAISS [27] index takes 0.09 ms (SigLIP 2) and 0.05 ms (DINO). These were measured on 500 fixed queries after warm-up, with the two stages timed separately, and are not end-to-end figures; they describe this implementation on this machine and are not compared with timings reported elsewhere.

## VII. DISCUSSION

*One ordering, not two mechanisms.* The results separate by backbone, but the separation is one of degree. For the encoder with no strong prior for printed labels, the views supply most of the accuracy and reproduce on public data what [4] and [7] reported on private data; for the text-supervised encoder that already retrieves labels accurately, the views still carry something the frozen embedding does not use, but it is worth about one point and is obtained only by an adaptation that leaves the encoder alone. These results favor preserving the pretrained features under the tested recipes. They do not isolate the effect of parameter count, because objectives, learning rates, and checkpoint budgets also differ. Geometric synthesis remains a substitute for a strong prior; as a supplement to one, it gives a small gain that depends on the recipe.

The comparison against the frozen model is a comparison within one experiment: all adapted models start from the same weights, train on the same classes, and are evaluated identically, so the frozen model is the no-adaptation control. Two limits apply. A contrastive objective closer to the model's pretraining, alternative ratios of original and synthesized images or additional real training photographs, adapters of other ranks, and heads of more than one layer were not tried, and we do not claim that no recipe would gain more than a point. Nor are the backbones measured against each other under a common recipe, since the DINO experiment keeps the recipe of [4]; they are compared only qualitatively.

*Why fine-tuning on the views loses accuracy.* The distance measurement, the views-only enrollment result, and the drift curves point in the same direction without proving a mechanism. In the SigLIP 2 space a geometrically synthesized view lies about three times as far from its source photograph as a real query sharing the dataset vintage identifier, and 22 points of accuracy are lost when the views replace the photograph as enrollment. A mismatch between the synthesized training views and real queries is one possible explanation for the declining validation accuracy, but these measurements do not establish that mechanism. The linear-head result fits the same account from the other side: when the encoder cannot move, the views can only reshape the metric over its features, and there the geometric views help and the 2D views do not, within the intervals.

The 2D pipeline does not fit this account: its images are the closest to the photographs, yet it loses the most under fine-tuning. Part of the explanation is likely the composition of the control itself. As used in [4] it includes horizontal flips and conversion to gray, so half of its views show mirrored text and some show no color; a text-supervised encoder trained to treat a mirrored label as the same class as the original is trained on a transformation that printed text never undergoes, whereas the geometric views, whatever their distance, preserve the reading direction. The flip-free control measures this. It retains 3.3 points more than the original 2D control after 4,000 updates and 1.6 more at the validation-chosen budget, so removing the flip operation improves this 2D pipeline but leaves most of its gap to the geometric pipelines, which remains 11.6 and 2.5 points; on the self-supervised backbone, removing the flip lowers accuracy by 2.2 points, so removing the flip has opposite effects under the two tested backbone–recipe combinations. The finding that geometric synthesis retains more accuracy than 2D augmentation under fine-tuning therefore does not depend on the flip.

*What the front end contributes.* The SAM front end's advantage over the edge-based one is, first, a matter of coverage: it synthesizes all six views for 99% of user photographs where the edge detector manages 43%, and the 2 × 2 of Table VI shows front-end gains under fixed-budget SigLIP 2 training as well as a smaller construction-associated difference. Whether it also produces better training views on the photographs where both front ends succeed is not settled. Under fixed-budget full fine-tuning the SAM-trained model is ahead on those sources, but it is ahead by a similar margin on the sources where the edge-based localization failed and the edge-based bank holds 2D images instead, so at that budget the difference is a property of the whole training bank acting through the shared encoder; at the validation-selected checkpoint and under the DINO recipe the two models cannot be told apart on the sources where both succeed, and neither could they in a 41-source development comparison. On the 100-class subset the SAM-based pipeline does separate from the edge-based one under both backbones, but this subset also changes the exposure per class and is not an independent replicate. The source of the difference was not isolated.

*When the construction matters.* The renderer isolates source tilt under controlled conditions: displacement rises more steeply for the frontal-assumption construction as tilt increases. Its nonzero frontal offset, caused by the crop and canvas adaptation, prevents a direct comparison of absolute geometric accuracy. On real photographs the recognition contrast depends on the recipe. The rim-estimating pipeline is higher after fixed-budget SigLIP 2 fine-tuning, whereas the linear-head contrast is unresolved. The pose-proxy analysis establishes no change in recognition contrast across tertiles; because the proxy is uncalibrated, it does not prove that tilted sources are absent. The results therefore identify a geometric sensitivity in simulation, while leaving its practical recognition benefit on verified tilted photographs open. Neither construction recovers rotation about the bottle's own axis from the rims alone.

*What remains.* The error audit highlights visually different query and enrollment labels, alongside confusions between similar designs and failures on different designs. Such cases warrant verification of both image-level appearance and the underlying vintage identity. Their presence does not establish an irreducible error rate. Evaluations on WineSensed-derived benchmarks would benefit from audited error breakdowns and verified producer and year labels. The metadata snapshot used here does not provide those labels across the benchmark, so producer-level matching and year-based analysis would require additional annotation.

*Limitations.* The benchmark inherits the limits of its source. Classes are vintage identifiers from crowd-sourced metadata, without human verification of identity at scale; queries and enrollment images may in some cases come from the same photographic session; the pretraining data of the public encoders may contain images from the same platform, which cannot be checked; the human audit covers 50 of 226 errors and was performed by one annotator; and the confidence intervals are per comparison. The pose proxy of Section V-A is uncalibrated and confounded with bottle shape, distance, and rim visibility, and its tertiles are cut on the benchmark sources themselves, although before any stratified result was seen.

The experiments have limits of their own. The self-supervised backbone is the DINO ViT-S/16 of [4], kept so that the earlier recipe could be reproduced; DINOv2-S, evaluated frozen on the development split (Table I), was not fine-tuned. Its development score is not directly comparable with the DINO benchmark score because the classes and queries differ. The full fine-tuning recipe uses a constant learning rate of $10^{-5}$ without warm-up, and smaller rates or schedules that might lose less were not tried; the training budgets are fixed and matched rather than converged; the linear head was run at one width and two learning rates and LoRA at one rank, and the two were not given equal budgets, the linear head's checkpoint grid running to 4,000 updates where LoRA's stops at 1,000; the two constructions differ in the scale and projection conventions of the label region (Section III-B), which the construction comparison does not control; and the fourth bank's DINO runs use a numerically stable evaluation of the loss that the other DINO runs did not need. The edge-based pipeline was run from its original implementation with crash fixes only, on photographs of lower resolution than those it was built for. The source-pose test of Table S2 gives the frontal-assumption construction rendered input through a crop and canvas adaptation that leaves it a 27 px error at zero tilt, so only its trend with tilt is interpretable. The second-stage and OCR results are development studies of particular implementations. No original-image-only adaptation control was run, so the gains over the frozen model cannot be assigned entirely to synthesized views; comparisons against the 2D-trained controls isolate a narrower difference between augmentation pipelines.

## VIII. CONCLUSION

On a public one-shot benchmark of 1,000 wine-label classes, frozen SigLIP 2-B reaches 94.7% top-1 from one enrollment photograph per class, while frozen DINO ViT-S/16 reaches 34.1%. Under the recovered DINO training recipe, geometric synthesis adds about 30 percentage points, compared with about 10 from the original 2D control. For SigLIP 2, a linear head over frozen features trained with the two SAM-localized geometric pipelines reaches about 96.0%. LoRA and validation-selected full fine-tuning do not show a clear gain within the reported intervals, and fixed-budget full fine-tuning loses accuracy. Maximum aggregation over multiple enrollment views changes accuracy by less than a quarter of a point. SAM localization raises complete synthesis coverage from 43% to 99%. Recognition differences between the cylinder constructions depend on the recipe and cannot be attributed to geometry alone because their rendering conventions differ. A controlled renderer shows different responses to source tilt, while the uncalibrated pose proxy establishes no corresponding trend in real-image recognition. The error audit identifies query–enrollment appearance mismatches without determining an accuracy ceiling. The public identifiers, splits, rankings, and evaluation tools support further tests of adaptation recipes and benchmarks with independently verified identities.

## DATA AVAILABILITY

Identifier lists, splits, hashes, synthesis parameters, per-query rankings, and evaluation code: evidence release v1 (2026-09-24), package SHA-256 36394cd3819efb389977aa5dc35845ae23235e734090903eb60a23eb63ec5f6c, at https://github.com/EdenHuang/winesensed-one-shot-benchmark. Images are available from the WineSensed release [12] and are not redistributed; the Hugging Face dataset card states CC BY-NC-ND 4.0 and the Technical University of Denmark (DTU) data record CC BY-NC 4.0, and users should consult the source.

## ACKNOWLEDGMENT

The author thanks Jen-Hui Chuang and Jenq-Neng Hwang, co-authors of the conference papers on which this work builds, for their guidance on that earlier work; Cheng-Jui Hung and Yu-Hao Chen, co-authors of those papers, whose experiment records and code enabled the recovery of the earlier pipelines; and the authors of WineSensed for releasing the dataset. Claude (Anthropic) assisted with drafting text in Sections I–VIII and figure captions. OpenAI Codex assisted with submission-stage editing, checks against saved results, and preparation of the public evidence release. The author reviewed the text, results, and interpretations and takes responsibility for the manuscript.

## REFERENCES

[1] X. Li, J. Yang, J. Ma, "CNN-SIFT consecutive searching and matching for wine label retrieval," in Intelligent Computing Theories and Application (ICIC 2019), Lecture Notes in Computer Science, vol. 11643. Cham, Switzerland: Springer, 2019, pp. 250–261, doi: 10.1007/978-3-030-26763-6_24.

[2] X. Li, J. Yang, J. Ma, "Large scale category-structured image retrieval for object identification through supervised learning of CNN and SURF-based matching," IEEE Access, vol. 8, pp. 57796–57809, 2020, doi: 10.1109/ACCESS.2020.2982560.

[3] X. Li, J. Ma, "Distributed search and fusion for wine label image retrieval," PeerJ Computer Science, vol. 8, Art. no. e1116, 2022, doi: 10.7717/peerj-cs.1116.

[4] Y.-C. Huang, H.-Y. Chen, C.-J. Hung, J.-H. Chuang, J.-N. Hwang, "Single-image driven 3D viewpoint training data augmentation for effective label recognition," in Pattern Recognition (ICPR 2024), Lecture Notes in Computer Science, vol. 15332. Cham, Switzerland: Springer, 2025, pp. 196–211, doi: 10.1007/978-3-031-78125-4_14.

[5] Y.-C. Huang, Y.-H. Chen, J.-H. Chuang, J.-N. Hwang, "HierarchicalWine: enhancing wine label recognition through advanced 3D data augmentation and hierarchical classification," in Proc. 2025 7th Int. Conf. Robotics and Computer Vision (ICRCV), 2025, pp. 101–108, doi: 10.1109/ICRCV67407.2025.11349206.

[6] A. Kirillov et al., "Segment anything," ICCV, 2023, pp. 4015–4026.

[7] X. Li, X. Zhang, Z. Cai, J. Ma, "On wine label image data augmentation through viewpoint based transformation," Journal of Signal Processing, vol. 38, no. 1, pp. 43–54, 2022 (in Chinese). DOI 10.16798/j.issn.1003-0530.2022.01.006.

[8] M. Caron et al., "Emerging properties in self-supervised vision transformers," in Proc. IEEE/CVF Int. Conf. Computer Vision (ICCV), 2021, pp. 9650–9660.

[9] M. Oquab et al., "DINOv2: learning robust visual features without supervision," TMLR, 2024.

[10] M. Tschannen, A. Gritsenko, X. Wang, et al., "SigLIP 2: multilingual vision-language encoders with improved semantic understanding, localization, and dense features," arXiv:2502.14786, 2025.

[11] G. Kordopatis-Zilos et al., "ILIAS: instance-level image retrieval at scale," in Proc. IEEE/CVF Conf. Computer Vision and Pattern Recognition (CVPR), 2025, pp. 14777–14787.

[12] T. Bender et al., "Learning to taste: a multimodal wine dataset," in Advances in Neural Information Processing Systems, vol. 36, Datasets and Benchmarks Track, 2023, doi: 10.52202/075280-0321.

[13] A. Angeli, L. Stacchio, L. Donatiello, A. Giacchè, G. Marfia, "Making paper labels smart for augmented wine recognition," The Visual Computer, vol. 40, no. 8, pp. 5519–5531, 2024, doi: 10.1007/s00371-023-03119-y.

[14] A. Zankevich, "How to unwrap wine labels programmatically," HackerNoon, Sep. 27, 2018. [Online]. Available: https://medium.com/hackernoon/how-to-unwrap-wine-labels-programmatically-31c8c62b30ce (accessed Sep. 26, 2026); FabImage, "Bottle flattening," FabImage Studio documentation. [Online]. Available: https://docs.fab-image.com/studio/examples/bottle_flattening.html (accessed Sep. 26, 2026).

[15] A. Tonioni, L. Di Stefano, "Domain invariant hierarchical embedding for grocery products recognition," Computer Vision and Image Understanding, vol. 182, pp. 81–92, 2019, doi: 10.1016/j.cviu.2019.03.005.

[16] P. Suma, G. Kordopatis-Zilos, A. Iscen, G. Tolias, "AMES: asymmetric and memory-efficient similarity estimation for instance-level retrieval," in Computer Vision – ECCV 2024. Cham, Switzerland: Springer, 2024, pp. 307–325, doi: 10.1007/978-3-031-73202-7_18.

[17] X. Bai et al., "Integrating scene text and visual appearance for fine-grained image classification," IEEE Access, vol. 6, pp. 66322–66335, 2018, doi: 10.1109/ACCESS.2018.2878899.

[18] A. Mafla et al., "Fine-grained image classification and retrieval by combining visual and locally pooled textual features," in Proc. 2020 IEEE Winter Conf. Applications of Computer Vision (WACV), 2020, pp. 2939–2948, doi: 10.1109/WACV45572.2020.9093373.

[19] A. Mafla et al., "Multi-modal reasoning graph for scene-text based fine-grained image classification and retrieval," in Proc. IEEE/CVF Winter Conf. Applications of Computer Vision (WACV), 2021, pp. 4023–4033.

[20] T. Pettersson, M. Riveiro, T. Löfström, "Multimodal fine-grained grocery product recognition using image and OCR text," Machine Vision and Applications, vol. 35, no. 4, art. 79, 2024. DOI 10.1007/s00138-024-01549-9.

[21] A. Zhang, J. Mazumder, K. Lomakin, "Bridging the catalog-to-real gap: scalable product recognition via multi-stage contrastive learning," arXiv:2607.09888, 2026.

[22] S. Liu et al., "Grounding DINO: marrying DINO with grounded pre-training for open-set object detection," in Computer Vision – ECCV 2024. Cham, Switzerland: Springer, 2024, pp. 38–55, doi: 10.1007/978-3-031-72970-6_3.

[23] D. G. Lowe, "Distinctive image features from scale-invariant keypoints," Int. J. Computer Vision, vol. 60, no. 2, pp. 91–110, Nov. 2004, doi: 10.1023/B:VISI.0000029664.99615.94.

[24] J. Sun et al., "LoFTR: detector-free local feature matching with transformers," in Proc. IEEE/CVF Conf. Computer Vision and Pattern Recognition (CVPR), 2021, pp. 8922–8931.

[25] D. DeTone, T. Malisiewicz, A. Rabinovich, "SuperPoint: self-supervised interest point detection and description," in Proc. 2018 IEEE/CVF Conf. Computer Vision and Pattern Recognition Workshops (CVPRW), 2018, doi: 10.1109/CVPRW.2018.00060.

[26] P. Lindenberger, P.-E. Sarlin, M. Pollefeys, "LightGlue: local feature matching at light speed," in Proc. 2023 IEEE/CVF Int. Conf. Computer Vision (ICCV), 2023, pp. 17581–17592, doi: 10.1109/ICCV51070.2023.01616.

[27] J. Johnson, M. Douze, H. Jégou, "Billion-scale similarity search with GPUs," IEEE Trans. Big Data, vol. 7, no. 3, pp. 535–547, Jul. 2021, doi: 10.1109/TBDATA.2019.2921572.

**YUEH-CHENG HUANG** received the Ph.D. degree from National Yang Ming Chiao Tung University, Hsinchu, Taiwan, in 2025. He was an R&D Deputy Manager with Accordance Technology and held internships with the University of Washington, the Industrial Technology Research Institute, and Microsoft Taiwan. Since August 2026, he has been an Assistant Professor with the Department of Computer Science and Information Engineering, National Dong Hwa University, Hualien, Taiwan, where he leads the Visual Intelligence Systems Laboratory. His research interests include computer vision, geometric vision, and the design of artificial intelligence algorithms for visual systems.